\PassOptionsToPackage{svgnames,dvipsnames}{xcolor} 
\documentclass[acmsmall,nonacm]{acmart}

\acmJournal{CSUR}
\acmYear{2026}\acmVolume{1}\acmNumber{1}\acmArticle{1}\acmMonth{7}
\setcopyright{acmlicensed}  
\usepackage{forest}
\useforestlibrary{edges}   
\usepackage{tikz}
\usetikzlibrary{arrows.meta,positioning,fit,backgrounds,shapes.geometric,calc}
\usepackage{pifont}

\newcommand{\yy}{\textcolor{ForestGreen}{\ding{52}}}      
\newcommand{\nn}{\textcolor{BrickRed}{\ding{56}}}         
\newcommand{\dm}{\textcolor{gray!65}{\textbf{--}}}         
\usepackage{longtable}
\usepackage{array}      
\usepackage{placeins}   

\newcommand{\tcode}{\textcolor{cat-code!55!black}{\textsf{\textbf{code}}}}
\newcommand{\tweights}{\textcolor{cat-vla!45!black}{\textsf{\textbf{weights}}}}
\newcommand{\treward}{\textcolor{cat-rew!55!black}{\textsf{\textbf{reward}}}}
\newcommand{\tskill}{\textcolor{cat-skill!45!black}{\textsf{\textbf{skill}}}}
\newcommand{\tpolicy}{\textcolor{cat-trans!55!black}{\textsf{\textbf{policy}}}}
\newcommand{\tbench}{\textcolor{cat-bench!45!black}{\textsf{\textbf{bench}}}}
\newcommand{\dreal}{\textcolor{ForestGreen}{\textsf{real}}}
\newcommand{\dsim}{\textcolor{RoyalBlue}{\textsf{sim}}}
\newcommand{\dgame}{\textcolor{Purple}{\textsf{game}}}
\newcommand{\dtext}{\textcolor{gray!70}{\textsf{text}}}
\newcommand{\flatent}{\textcolor{secA!45!black}{\textsf{latent-RL}}}
\newcommand{\fspace}{\textcolor{secB!45!black}{\textsf{skill-space}}}
\newcommand{\fcodelib}{\textcolor{secF!45!black}{\textsf{code-lib}}}
\newcommand{\fagent}{\textcolor{secG!45!black}{\textsf{agent-lib}}}
\newcommand{\pp}{\textcolor{orange!90!black}{\textbf{\textasciitilde}}}  

\definecolor{hidden-draw}{RGB}{40,40,40}
\definecolor{tx-root}{HTML}{D1D5DB}

\definecolor{cat-code}{HTML}{93C5FD}\definecolor{leaf-code}{HTML}{DBEAFE}   
\definecolor{cat-vla}{HTML}{5EEAD4}\definecolor{leaf-vla}{HTML}{CCFBF1}     
\definecolor{cat-rew}{HTML}{FDBA74}\definecolor{leaf-rew}{HTML}{FEEBC8}     
\definecolor{cat-skill}{HTML}{C4B5FD}\definecolor{leaf-skill}{HTML}{EDE9FE} 
\definecolor{cat-trans}{HTML}{F9A8D4}\definecolor{leaf-trans}{HTML}{FCE7F3} 
\definecolor{cat-bench}{HTML}{86EFAC}\definecolor{leaf-bench}{HTML}{DCFCE7} 
\definecolor{barneutral}{HTML}{64748B} 

\definecolor{secA}{HTML}{93C5FD}\definecolor{secAl}{HTML}{DBEAFE} 
\definecolor{secB}{HTML}{5EEAD4}\definecolor{secBl}{HTML}{CCFBF1} 
\definecolor{secC}{HTML}{86EFAC}\definecolor{secCl}{HTML}{DCFCE7} 
\definecolor{secD}{HTML}{FCD34D}\definecolor{secDl}{HTML}{FEF3C7} 
\definecolor{secE}{HTML}{FCA5A5}\definecolor{secEl}{HTML}{FEE2E2} 
\definecolor{secF}{HTML}{C4B5FD}\definecolor{secFl}{HTML}{EDE9FE} 
\definecolor{secG}{HTML}{F9A8D4}\definecolor{secGl}{HTML}{FCE7F3} 

\forestset{
  default preamble={
    forked edges,
    for tree={
      grow=east,
      reversed=true,
      anchor=base west,
      parent anchor=east,
      child anchor=west,
      base=center,
      font=\large,
      rectangle,
      draw=hidden-draw,
      rounded corners,
      align=center,
      text centered,
      minimum width=5em,
      edge+={darkgray, line width=1pt},
      s sep=3pt,
      inner xsep=2pt,
      inner ysep=3pt,
      line width=0.8pt,
      ver/.style={rotate=90, child anchor=north, parent anchor=south, anchor=center},
      if level=1{text width=19em, font=\normalsize}{},
      if level=2{text width=33em, font=\normalsize}{},
    },
  },
}

\begin{document}

\title{Weights or Skills?}
\subtitle{A Survey of Robot-Learning Techniques: from Action-Predicting
  Weights to Robots that Write their Own Skills}

\author{Gaytri Jena}
\authornote{The authors contributed to this work independently of their roles and employment.}
\affiliation{\institution{UC Berkeley}\city{Berkeley}\state{CA}\country{USA}}
\author{Kapil Wanaskar}
\affiliation{\institution{San Jose State University}\country{USA}}
\author{Vinija Jain}
\affiliation{\institution{Meta}\country{USA}}
\author{Aman Chadha}
\affiliation{\institution{Apple}\country{USA}}
\author{Vasu Sharma}
\affiliation{\institution{PocketFM}\country{USA}}
\author{Amitava Das}
\affiliation{\institution{Pragya Lab, BITS Pilani Goa}\country{India}}

\makeatletter
\if@ACM@anonymous\else
  \renewcommand{\shortauthors}{Jena et al.}
\fi
\makeatother

\begin{abstract}
Robot learning is splitting into two bets: policies that bake competence into \emph{frozen
weights} (vision-language-action, or VLA, models), and agents that write and refine their own
\emph{executable skills} as code. This survey organises the field around that axis of
\textbf{weights versus skills}. Its central analytical contribution is a deep-dive that arranges
code-as-policy methods by their \emph{degree of self-improvement}, from zero-shot program
synthesis, through closed-loop self-repair and persistent skill memory, to the sparsely
populated cell in which execution feedback, skill memory, and evolutionary search combine into
one open-ended loop; only a few very recent systems (for example ASPIRE, ENPIRE, and RoboClaw)
occupy that cell. We map the complementary ``skills'' pole, from unsupervised
reinforcement-learning skill discovery to large-language-model skill libraries, and show that
the word ``skill'' is used in at least five distinct senses, of which only the code sense
self-improves without gradient updates. We then connect the taxonomy to the emerging
\emph{skill economy}: commercial robot-skill marketplaces now distribute one-tap skills across
robots but ship only static playback, which surfaces open problems of adaptation,
cross-embodiment portability, provenance, safety verification, composition, and
standardisation. This is a deliberately \emph{focused} survey. Rather than cataloguing the
field exhaustively, it examines 77 representative systems across six technique families through
one taxonomy and a set of contrast tables, and it supplies operational definitions of the
self-improvement mechanisms together with a statement of what each family cannot do.
\end{abstract}



\maketitle

\section{Introduction}
\label{sec:intro}

Two paradigms now dominate robot learning. \emph{Vision-language-action} (VLA)
models predict actions from frozen weights; \emph{code-as-policy} agents instead
write executable programs and improve them from experience~\citep{codeaspolicies,aspire}.
This survey is organised around that fault line. Figure~\ref{fig:corpus_collage} poses the question
with real systems: frozen-weight VLA policies on one side, executable and self-improving
code-as-policy agents on the other.

\input{figures/fig_corpus_collage}

\paragraph{Why now: the skill stack.}
The app store for robot skills has arrived; Unitree UniStore ships one-tap,
cross-model motion downloads~\citep{unistore2026}. But every skill in it is
\emph{static playback}: the least-capable point in the taxonomy. The skill stack has
three layers: \textbf{(1) Build} (how skills are created, \S\ref{sec:taxonomy}--\S\ref{sec:techniques}),
\textbf{(2) Distribute} (marketplaces and hubs), and \textbf{(3) Adapt} (on-device
self-improvement, \S\ref{sec:code}). UniStore is layer~2 over static layer-1 skills;
the promised value, ``adjust grip per fruit so nothing bruises'', lives in layer~3,
which no shipped skill does today.

\paragraph{Scope and contributions.}
This is a \emph{focused} survey rather than an exhaustive census. We examine 77 systems chosen,
by the explicit placement criteria of \S\ref{sec:method}, to populate a single analytical
taxonomy, and we exclude pure perception, navigation, and
locomotion work except where it bears directly on how a manipulation skill is authored and
improved. Our contributions are: (i) a taxonomy organised by \emph{weights vs.\ skills}
(\S\ref{sec:taxonomy}); (ii) a deep-dive that arranges code-as-policy by \emph{degree of
self-improvement}, with operational definitions of the feedback, memory, and search mechanisms
and necessary-and-sufficient conditions for each rung (\S\ref{sec:code}); (iii) a map of the
``skills'' pole together with an analysis of the five senses in which ``skill'' is used
(\S\ref{sec:skill}); (iv) a summary of the characteristic strengths and limitations of each
technique family; and (v) an open-problems agenda for the emerging skill economy
(\S\ref{sec:open}). To situate these systems within the wider field,
Appendix~\ref{sec:landscape} (Table~\ref{tab:landscape}) catalogues 225 further representative
works, compared on six axes and grouped by area, bringing the survey's total coverage to more than
three hundred works.

\paragraph{Related surveys.}
Table~\ref{tab:survey_compare} contrasts the topical coverage of the closest surveys with
ours, and Figure~\ref{fig:surveys} places them on a timeline. Within \emph{ACM Computing
Surveys} the nearest are on world models, embodied intelligence, and language-model
navigation, and beyond it a cluster covers vision-language-action models and foundation models
for manipulation; but none organises code-as-policy by degree of self-improvement or names
the \S3.1.5 cell.
\begin{table}[t]
\centering
\footnotesize
\renewcommand{\arraystretch}{1.25}
\begin{tabular}{@{}p{0.24\textwidth}c c *{6}{c}@{}}
\toprule
\textbf{Survey} & \textbf{Venue} & \textbf{Yr}
 & \tweights{} & \tcode{} & \tskill{} & \treward{}
 & \textcolor{secE!55!black}{\textbf{Self-impr.}} & \textcolor{cat-bench!45!black}{\textbf{Economy}}\\
\midrule
World Models~\cite{survey_worldmodels} & CSUR & 2025 & \pp & \nn & \nn & \nn & \nn & \nn\\
Embodied Intelligence~\cite{survey_embodiedintel}& CSUR & 2025 & \pp & \nn & \pp & \nn & \nn & \nn\\
Navigation via FLM~\cite{survey_navigation} & CSUR & 2026 & \pp & \pp & \nn & \nn & \nn & \nn\\
VLA for Embodied AI~\cite{survey_vla} & arXiv & 2024 & \yy & \nn & \nn & \nn & \nn & \nn\\
Large-VLM VLA~\cite{survey_vlmvla} & arXiv & 2025 & \yy & \nn & \nn & \pp & \nn & \nn\\
Embodied Learning~\cite{survey_embodiedlearn} & MIR & 2025 & \pp & \nn & \pp & \pp & \nn & \nn\\
FM for Manipulation~\cite{survey_fmrobot} & arXiv & 2024 & \yy & \pp & \pp & \pp & \nn & \nn\\
\midrule
\textbf{Ours (\emph{Weights or Skills?})} & \dm & 2026 & \yy & \yy & \yy & \yy & \yy & \yy\\
\bottomrule
\end{tabular}
\caption{Topical coverage of the closest related surveys versus ours.
\yy\ covered, \pp\ partial/mentioned, \nn\ not covered. Columns: \tweights{} (VLA/weights),
\tcode{} (code-as-policy), \tskill{} (skill libraries), \treward{} (reward synthesis),
\textbf{Self-impr.}\ (organised by \emph{degree of self-improvement}, incl.\ the \S3.1.5 cell),
\textbf{Economy} (the robot skill marketplace). Only this survey covers the last two axes.
CSUR = ACM Computing Surveys; MIR = Machine Intelligence Research.}
\label{tab:survey_compare}
\end{table}

\begin{figure*}[p]
\centering
\resizebox{!}{0.88\textheight}{%
\begin{forest}
  for tree={align=center},
  where n children=0{font=\footnotesize, align=left}{font=\normalsize},
  where level=1{font=\large}{},
  where level=0{font=\Large}{},
[\textbf{Robot-learning}\\\textbf{techniques}\\\textbf{for Physical AI}, fill=tx-root, text width=8.5em
  [\textbf{\S3.1}\\\textbf{Code-as-policy}, fill=cat-code, text width=9em
    [\textbf{\S3.1.1 Zero-shot}\\\textbf{synthesis}, fill=secA, text width=9.5em
      [{\textbf{Code-as-Policies}: LLM programs for embodied control~\citep{codeaspolicies}\\\textbf{ProgPrompt}: situated robot task plans via LLMs~\citep{progprompt}\\\textbf{VoxPoser}: composable 3D value maps~\citep{voxposer}\\\textbf{Instruct2Act}: multimodal instructions to actions~\citep{instruct2act}\\\textbf{ChatGPT for Robotics}: design principles \& abilities~\citep{chatgptrobotics}\\\textbf{TidyBot}: personalised robot assistance~\citep{tidybot}\\\textbf{RoboCodeX}: multimodal code for behavior synthesis~\citep{robocodex}\\\textbf{RoboScript}: code for free-form manipulation~\citep{roboscript}\\\textbf{Text2Motion}: instructions to feasible plans~\citep{text2motion}\\\textbf{Demo2Code}: demonstrations to synthesized code~\citep{demo2code}\\\textbf{Statler}: state-maintaining LLMs for reasoning~\citep{statler}\\\textbf{Prompt2Walk}: prompt a robot to walk with LLMs~\citep{prompt2walk}\\\textbf{RoboPro}: video-instructed policy code generation~\citep{robopro}}, fill=secAl, text width=33em, align=left]
    ]
    [\textbf{\S3.1.2 Closed-loop}\\\textbf{self-repair}, fill=secB, text width=9.5em
      [{\textbf{Inner Monologue}: embodied reasoning via planning~\citep{innermonologue}\\\textbf{DoReMi}: recovering plan-execution misalignment~\citep{doremi}\\\textbf{REFLECT}: summarizing experiences for failure explanation~\citep{reflect}\\\textbf{Code-as-Monitor}: constraint-aware visual programming~\citep{codeasmonitor}\\\textbf{AHA}: VLM reasoning over manipulation failures~\citep{aha}\\\textbf{Introspective Planning}: uncertainty vs.\ task ambiguity~\citep{introspective}}, fill=secBl, text width=33em, align=left]
    ]
    [\textbf{\S3.1.3 Skill-library}\\\textbf{accumulation}, fill=secC, text width=9.5em
      [{\textbf{RoboCoder}: from basic skills to general tasks~\citep{robocoder}\\\textbf{DROC}: distilling knowledge via language corrections~\citep{droc}}, fill=secCl, text width=33em, align=left]
    ]
    [\textbf{\S3.1.4 Evolutionary}\\\textbf{search}, fill=secD, text width=9.5em
      [{\textbf{CaP-X}: benchmarking \& improving coding agents~\citep{capx}\\\textbf{RoboEvolve}: co-evolving planner-simulator~\citep{roboevolve}\\\textbf{GEAR}: policies via LLM evolutionary search~\citep{codeevolution}}, fill=secDl, text width=33em, align=left]
    ]
    [\textbf{\S3.1.5 Full}\\\textbf{self-improving loop}, fill=secE, text width=9.5em
      [{\textbf{ASPIRE}: agentic skills discovery for robotics~\citep{aspire}\\\textbf{ENPIRE}: real-world policy self-improvement~\citep{enpire}\\\textbf{RoboClaw}: scalable long-horizon robotic tasks~\citep{roboclaw}}, fill=secEl, text width=33em, align=left]
    ]
  ]
  [\textbf{\S3.2 End-to-end}\\\textbf{VLA}, fill=cat-vla, text width=9em
    [{\textbf{RT-1}: robotics transformer for real-world control~\citep{rt1}\\\textbf{RT-2}: VLA models transfer web knowledge~\citep{rt2}\\\textbf{Octo}: open-source generalist robot policy~\citep{octo}\\\textbf{OpenVLA}: open-source VLA model~\citep{openvla}\\\textbf{RoboFlamingo}: VLM foundation models as imitators~\citep{roboflamingo}\\\textbf{CogACT}: synergizing cognition and action~\citep{cogact}\\\textbf{SpatialVLA}: spatial representations for VLA~\citep{spatialvla}\\$\boldsymbol{\pi_0}$: VLA flow model for general control~\citep{pi0}\\$\boldsymbol{\pi_{0.5}}$: VLA with open-world generalisation~\citep{pi05}\\\textbf{GR00T N1}: foundation model for humanoid robots~\citep{groot}\\\textbf{Gemini Robotics}: bringing AI into the physical world~\citep{geminirobotics}}, fill=leaf-vla, text width=33em, align=left]
  ]
  [\textbf{\S3.3 Reward}\\\textbf{synthesis}, fill=cat-rew, text width=9em
    [{\textbf{Eureka}: human-level reward design via LLMs~\citep{eureka}\\\textbf{DrEureka}: LLM-guided sim-to-real transfer~\citep{dreureka}\\\textbf{Text2Reward}: reward shaping with LLMs~\citep{text2reward}\\\textbf{Language-to-Rewards}: for robotic skill synthesis~\citep{l2r}\\\textbf{Eurekaverse}: environment curriculum generation~\citep{eurekaverse}\\\textbf{RoboGen}: automated learning via generative sim~\citep{robogen}\\\textbf{Auto MC-Reward}: automated dense reward design~\citep{automcreward}}, fill=leaf-rew, text width=33em, align=left]
  ]
  [\textbf{\S3.4 Skill}\\\textbf{libraries}, fill=cat-skill, text width=9em
    [\textbf{\S3.4.1 Unsupervised}\\\textbf{latent-RL}, fill=cat-skill, text width=9.5em
      [{\textbf{DIAYN}: learning skills without a reward function~\citep{diayn}\\\textbf{DADS}: dynamics-aware discovery of skills~\citep{dads}\\\textbf{LSD}: Lipschitz-constrained skill discovery~\citep{lsd}\\\textbf{CIC}: contrastive intrinsic control~\citep{cic}\\\textbf{METRA}: scalable unsupervised RL, metric-aware~\citep{metra}}, fill=leaf-skill, text width=33em, align=left]
    ]
    [\textbf{\S3.4.2 Skill-space}\\\textbf{\& hierarchical RL}, fill=cat-skill, text width=9.5em
      [{\textbf{SPiRL}: accelerating RL with learned skill priors~\citep{spirl}\\\textbf{OPAL}: offline primitive discovery~\citep{opal}\\\textbf{PARROT}: data-driven behavioral priors~\citep{parrot}\\\textbf{SkiMo}: skill-based model-based RL~\citep{skimo}}, fill=leaf-skill, text width=33em, align=left]
    ]
    [\textbf{\S3.4.3 LLM \&}\\\textbf{code libraries}, fill=cat-skill, text width=9.5em
      [{\textbf{Voyager}: open-ended embodied agent with LLMs~\citep{voyager}\\\textbf{LOTUS}: continual imitation via skill discovery~\citep{lotus}\\\textbf{LRLL}: bootstrapping composable skills~\citep{lrll}\\\textbf{BOSS}: learning new tasks with LLM guidance~\citep{boss}\\\textbf{SPRINT}: pre-training via instruction relabeling~\citep{sprint}\\\textbf{Uni-Skill}: self-evolving skill repository~\citep{uniskill}\\\textbf{SkillFlow}: lifelong skill discovery \& evolution~\citep{skillflow}}, fill=leaf-skill, text width=33em, align=left]
    ]
    [\textbf{\S3.4.4 Open-world}\\\textbf{LLM agents}, fill=cat-skill, text width=9.5em
      [{\textbf{GITM}: capable agents with text-based memory~\citep{gitm}\\\textbf{JARVIS-1}: memory-augmented multimodal agents~\citep{jarvis1}\\\textbf{ExpeL}: LLM agents are experiential learners~\citep{expel}\\\textbf{Optimus-1}: hybrid multimodal memory agents~\citep{optimus1}\\\textbf{Odyssey}: Minecraft agents with open-world skills~\citep{odyssey}}, fill=leaf-skill, text width=33em, align=left]
    ]
  ]
  [\textbf{\S3.5 Sim-to-real}\\\textbf{\& transfer}, fill=cat-trans, text width=9em
    [{\textbf{Open X-Embodiment}: datasets and RT-X models~\citep{openx}\\\textbf{CrossFormer}: one policy for many embodiments~\citep{crossformer}\\\textbf{RoboCat}: self-improving generalist agent~\citep{robocat}\\\textbf{Mirage}: zero-shot transfer via cross-painting~\citep{mirage}}, fill=leaf-trans, text width=33em, align=left]
  ]
  [\textbf{\S3.6 Benchmarks}\\\textbf{\& simulators}, fill=cat-bench, text width=9em
    [{\textbf{LIBERO}: knowledge transfer for lifelong learning~\citep{libero}\\\textbf{robosuite}: modular simulation framework~\citep{robosuite}\\\textbf{BEHAVIOR-1K}: 1000 everyday household activities~\citep{behavior1k}\\\textbf{Meta-World}: benchmark for multi-task \& meta RL~\citep{metaworld}\\\textbf{RLBench}: robot learning benchmark \& environment~\citep{rlbench}\\\textbf{ManiSkill2}: unified benchmark for manipulation~\citep{maniskill2}\\\textbf{CALVIN}: language-conditioned long-horizon tasks~\citep{calvin}}, fill=leaf-bench, text width=33em, align=left]
  ]
]
\end{forest}}
\caption{\textbf{The weights-versus-skills taxonomy of robot learning} (all 77 systems). Each
sub-family is one cell listing its systems, each with a trimmed form of the paper's own title.
Left fork = inspectable \emph{code / skills} (\S3.1, \S3.4); the \S3.1 cells are shaded by
\emph{degree of self-improvement}, from \colorbox{secAl}{\scriptsize zero-shot} to
\colorbox{secEl}{\scriptsize the full feedback+memory+search loop}. Full per-system detail:
Tables~\ref{tab:compare_main}--\ref{tab:compare_34}.}
\Description{A tree diagram organising 77 robot-learning systems under the root ``robot-learning
techniques for physical AI'' into six branches: code-as-policy (with five self-improvement rungs from
zero-shot synthesis to a full feedback-plus-memory-plus-search loop), end-to-end vision-language-action
models, reward synthesis, skill libraries, sim-to-real transfer, and benchmarks; each branch lists its
constituent systems.}
\label{fig:taxonomy_main}
\end{figure*}
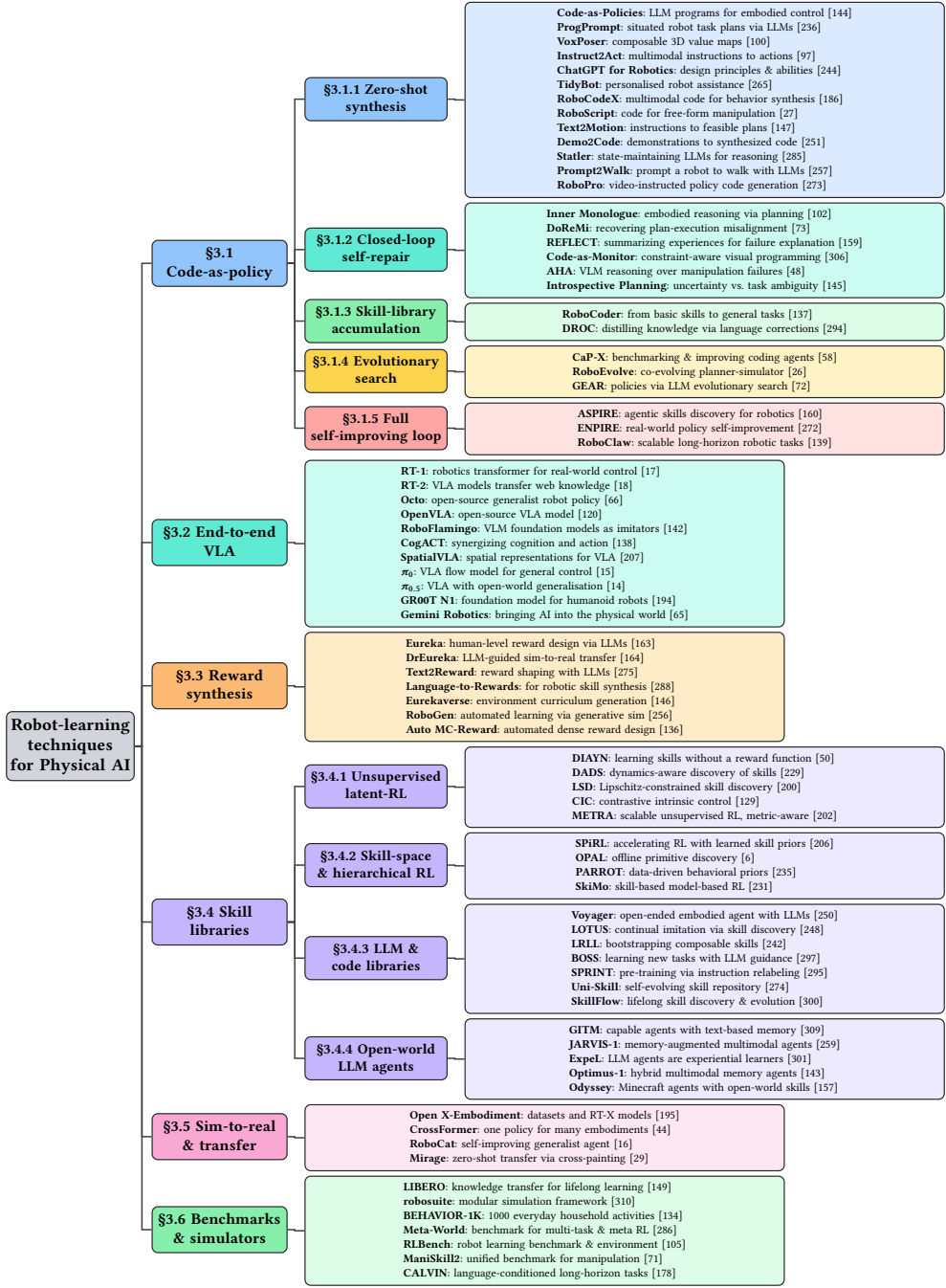

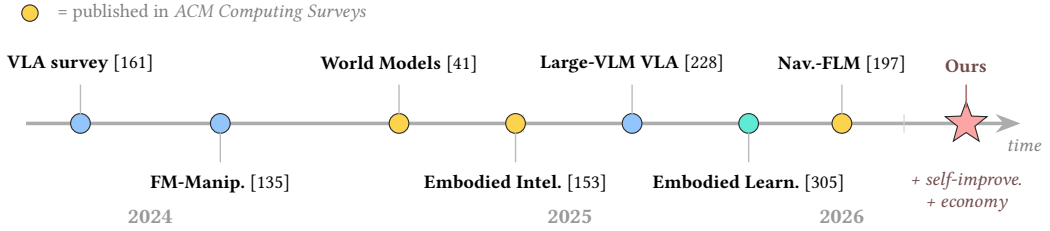
\begin{figure}[t]
\centering
\resizebox{\columnwidth}{!}{%
\begin{tikzpicture}[
 abv/.style={font=\scriptsize, anchor=south, align=center},
 blw/.style={font=\scriptsize, anchor=north, align=center},
 stem/.style={gray!55, line width=0.5pt},
]
 \draw[-{Stealth[length=3.2mm]}, line width=1.3pt, gray!65] (-0.4,0) -- (12.4,0);
 \node[font=\scriptsize\itshape, gray, anchor=west] at (12.1,-0.28) {time};
 \foreach \x in {0.3,6,10.9}{\draw[gray!35] (\x,-0.12) -- (\x,0.12);}
 \node[font=\footnotesize\bfseries,gray!80] at (1.2,-1.2) {2024};
 \node[font=\footnotesize\bfseries,gray!80] at (6.6,-1.2) {2025};
 \node[font=\footnotesize\bfseries,gray!80] at (10.1,-1.2) {2026};
 \filldraw[fill=secD,draw=hidden-draw] (-0.35,1.42) circle (3pt);
 \node[font=\scriptsize,gray,anchor=west] at (-0.12,1.42) {= published in \emph{ACM Computing Surveys}};

 \filldraw[fill=secA,draw=hidden-draw] (0.3,0) circle (3.6pt);
 \draw[stem] (0.3,0.12) -- (0.3,0.5); \node[abv] at (0.3,0.53) {\textbf{VLA survey}~\cite{survey_vla}};

 \filldraw[fill=secA,draw=hidden-draw] (2.1,0) circle (3.6pt);
 \draw[stem] (2.1,-0.12) -- (2.1,-0.5); \node[blw] at (2.1,-0.53) {\textbf{FM-Manip.}~\cite{survey_fmrobot}};

 \filldraw[fill=secD,draw=hidden-draw] (4.4,0) circle (3.6pt);
 \draw[stem] (4.4,0.12) -- (4.4,0.5); \node[abv] at (4.4,0.53) {\textbf{World Models}~\cite{survey_worldmodels}};

 \filldraw[fill=secD,draw=hidden-draw] (5.9,0) circle (3.6pt);
 \draw[stem] (5.9,-0.12) -- (5.9,-0.5); \node[blw] at (5.9,-0.53) {\textbf{Embodied Intel.}~\cite{survey_embodiedintel}};

 \filldraw[fill=secA,draw=hidden-draw] (7.4,0) circle (3.6pt);
 \draw[stem] (7.4,0.12) -- (7.4,0.5); \node[abv] at (7.4,0.53) {\textbf{Large-VLM VLA}~\cite{survey_vlmvla}};

 \filldraw[fill=secB,draw=hidden-draw] (8.9,0) circle (3.6pt);
 \draw[stem] (8.9,-0.12) -- (8.9,-0.5); \node[blw] at (8.9,-0.53) {\textbf{Embodied Learn.}~\cite{survey_embodiedlearn}};

 \filldraw[fill=secD,draw=hidden-draw] (10.1,0) circle (3.6pt);
 \draw[stem] (10.1,0.12) -- (10.1,0.5); \node[abv] at (10.1,0.53) {\textbf{Nav.-FLM}~\cite{survey_navigation}};

 \node[star,star points=5,star point ratio=2.4,fill=secE,draw=hidden-draw,minimum size=16pt,inner sep=0] at (11.7,0) {};
 \draw[secE!70!black,line width=0.6pt] (11.7,0.2) -- (11.7,0.5);
 \node[abv,text=secE!45!black] at (11.7,0.53) {\textbf{Ours}};
 \node[blw,text=secE!45!black,font=\scriptsize\itshape] at (11.7,-0.5) {+ self-improve.\\+ economy};
\end{tikzpicture}}
\caption{The closest related surveys as a timeline (2024--2026);
\textcolor{secD!55!black}{amber} markers denote surveys published in \emph{ACM Computing
Surveys} (CSUR). Prior surveys cover VLA models, foundation models, world models, and navigation,
but none organises code-as-policy by \emph{degree of self-improvement}, the weights--skills axis of
Figure~\ref{fig:corpus_collage}. \textbf{Ours}
($\star$, 2026) is organised around the self-improvement axis (\S3.1.5) and the skill economy;
per-topic coverage is in Table~\ref{tab:survey_compare}.}
\Description{A horizontal timeline of related surveys from 2024 to 2026, with amber markers for surveys
published in ACM Computing Surveys and other markers for arXiv surveys covering vision-language-action
models, foundation models, world models, and navigation; a red star at 2026 marks the present survey.}
\label{fig:surveys}
\end{figure}
\FloatBarrier

\section{A Taxonomy of Robot-Learning Techniques}
\label{sec:taxonomy}

Figure~\ref{fig:taxonomy_main} organises the field into six branches. The dividing
question is \emph{what ships}: frozen weights (\S\ref{sec:vla}) or executable skills
(\S\ref{sec:code}). We tag mechanisms by \textbf{F}eedback, \textbf{S}earch, and
\textbf{M}emory throughout.

Table~\ref{tab:compare_main} compares all systems in the map on what each \emph{ships}, where
it is evaluated, and whether it improves from its own experience.
{\footnotesize
\begin{longtable}{@{}p{0.30\textwidth}cll c c@{}}
\caption{Detailed comparison of a representative 47 of the 77 systems in the taxonomy
(Figure~\ref{fig:taxonomy_main}), grouped by family; the \S3.1 and \S3.4 branches are tabulated in
full in Tables~\ref{tab:compare_31} and~\ref{tab:compare_34}. \textbf{Ships}: \tcode/\tweights/\treward/\tskill/\tpolicy/\tbench.
\textbf{Eval}: \dreal/\dsim/\dgame. \textbf{Self-impr.}: \yy\ learns from its own experience,
\pp\ partially, \nn\ no, \dm\ n/a.}\label{tab:compare_main}\\
\toprule
\textbf{System} & \textbf{Year} & \textbf{Venue} & \textbf{Ships} & \textbf{Eval} & \textbf{Self-impr.}\\
\midrule
\endfirsthead
\multicolumn{6}{@{}l}{\footnotesize\emph{Table~\ref{tab:compare_main} (continued)}}\\
\toprule
\textbf{System} & \textbf{Year} & \textbf{Venue} & \textbf{Ships} & \textbf{Eval} & \textbf{Self-impr.}\\
\midrule
\endhead
\bottomrule
\endlastfoot
\multicolumn{6}{@{}l}{\textbf{\S3.1 Code-as-policy for robots} \hfill \tcode}\\
\textbf{Code-as-Policies}~\cite{codeaspolicies} & 2023 & ICRA & \tcode & \dreal & \nn\\
\textbf{ProgPrompt}~\cite{progprompt} & 2023 & ICRA & \tcode & \dreal & \nn\\
\textbf{VoxPoser}~\cite{voxposer} & 2023 & CoRL & \tcode & \dreal & \nn\\
\textbf{RoboCodeX}~\cite{robocodex} & 2024 & ICML & \tcode & \dsim & \nn\\
\textbf{Instruct2Act}~\cite{instruct2act} & 2023 & arXiv & \tcode & \dsim & \nn\\
\textbf{RoboPro}~\cite{robopro} & 2025 & arXiv & \tcode & \dreal & \nn\\
\textbf{RoboCoder}~\cite{robocoder} & 2024 & arXiv & \tcode & \dsim & \nn\\
\textbf{CaP-X}~\cite{capx} & 2026 & arXiv & \tcode & \dsim & \yy\\
\textbf{RoboClaw}~\cite{roboclaw} & 2026 & arXiv & \tcode & \dreal & \yy\\
\textbf{ASPIRE}~\cite{aspire} & 2026 & arXiv & \tcode & \dreal & \yy\\
\textbf{ENPIRE}~\cite{enpire} & 2026 & arXiv & \tcode & \dreal & \yy\\
\midrule
\multicolumn{6}{@{}l}{\textbf{\S3.2 End-to-end VLA / generalist policies} \hfill \tweights}\\
\textbf{RT-1}~\cite{rt1} & 2023 & RSS & \tweights & \dreal & \nn\\
\textbf{RT-2}~\cite{rt2} & 2023 & CoRL & \tweights & \dreal & \nn\\
\textbf{Octo}~\cite{octo} & 2024 & RSS & \tweights & \dreal & \nn\\
\textbf{OpenVLA}~\cite{openvla} & 2024 & CoRL & \tweights & \dreal & \nn\\
\textbf{RoboFlamingo}~\cite{roboflamingo} & 2024 & ICLR & \tweights & \dsim & \nn\\
\textbf{CogACT}~\cite{cogact} & 2024 & arXiv & \tweights & \dreal & \nn\\
\textbf{SpatialVLA}~\cite{spatialvla} & 2025 & arXiv & \tweights & \dreal & \nn\\
$\boldsymbol{\pi_0}$~\cite{pi0} & 2024 & arXiv & \tweights & \dreal & \nn\\
$\boldsymbol{\pi_{0.5}}$~\cite{pi05} & 2025 & arXiv & \tweights & \dreal & \nn\\
\textbf{GR00T N1}~\cite{groot} & 2025 & arXiv & \tweights & \dreal & \nn\\
\textbf{Gemini Robotics}~\cite{geminirobotics} & 2025 & arXiv & \tweights & \dreal & \nn\\
\midrule
\multicolumn{6}{@{}l}{\textbf{\S3.3 LLM-authored rewards / curricula} \hfill \treward}\\
\textbf{Eureka}~\cite{eureka} & 2024 & ICLR & \treward & \dsim & \yy\\
\textbf{DrEureka}~\cite{dreureka} & 2024 & RSS & \treward & \dreal & \yy\\
\textbf{Text2Reward}~\cite{text2reward} & 2024 & ICLR & \treward & \dsim & \pp\\
\textbf{Language-to-Rewards}~\cite{l2r} & 2023 & CoRL & \treward & \dreal & \pp\\
\textbf{Eurekaverse}~\cite{eurekaverse} & 2024 & CoRL & \treward & \dsim & \yy\\
\textbf{RoboGen}~\cite{robogen} & 2024 & ICML & \treward & \dsim & \yy\\
\textbf{Auto MC-Reward}~\cite{automcreward} & 2024 & CVPR & \treward & \dgame & \yy\\
\midrule
\multicolumn{6}{@{}l}{\textbf{\S3.4 Robot skill libraries \& lifelong learning} \hfill \tskill}\\
\textbf{Voyager}~\cite{voyager} & 2023 & arXiv & \tskill & \dgame & \yy\\
\textbf{LOTUS}~\cite{lotus} & 2024 & ICRA & \tskill & \dsim & \yy\\
\textbf{LRLL}~\cite{lrll} & 2024 & ICRA & \tskill & \dsim & \yy\\
\textbf{BOSS}~\cite{boss} & 2023 & CoRL & \tskill & \dsim & \yy\\
\textbf{SPRINT}~\cite{sprint} & 2024 & ICRA & \tskill & \dsim & \pp\\
\textbf{Uni-Skill}~\cite{uniskill} & 2026 & ICRA & \tskill & \dreal & \yy\\
\textbf{SkillFlow}~\cite{skillflow} & 2026 & arXiv & \tskill & \dtext & \yy\\
\midrule
\multicolumn{6}{@{}l}{\textbf{\S3.5 Sim-to-real \& cross-embodiment transfer} \hfill \tpolicy}\\
\textbf{Open X-Embodiment}~\cite{openx} & 2024 & ICRA & \tpolicy & \dreal & \nn\\
\textbf{CrossFormer}~\cite{crossformer} & 2024 & CoRL & \tpolicy & \dreal & \nn\\
\textbf{RoboCat}~\cite{robocat} & 2023 & TMLR & \tpolicy & \dreal & \yy\\
\textbf{Mirage}~\cite{mirage} & 2024 & RSS & \tpolicy & \dreal & \nn\\
\midrule
\multicolumn{6}{@{}l}{\textbf{\S3.6 Embodied benchmarks \& simulators} \hfill \tbench}\\
\textbf{LIBERO}~\cite{libero} & 2023 & NeurIPS & \tbench & \dsim & \dm\\
\textbf{Robosuite}~\cite{robosuite} & 2020 & arXiv & \tbench & \dsim & \dm\\
\textbf{BEHAVIOR-1K}~\cite{behavior1k} & 2024 & arXiv & \tbench & \dsim & \dm\\
\textbf{Meta-World}~\cite{metaworld} & 2019 & CoRL & \tbench & \dsim & \dm\\
\textbf{RLBench}~\cite{rlbench} & 2020 & RA-L & \tbench & \dsim & \dm\\
\textbf{ManiSkill2}~\cite{maniskill2} & 2023 & ICLR & \tbench & \dsim & \dm\\
\textbf{CALVIN}~\cite{calvin} & 2022 & RA-L & \tbench & \dsim & \dm\\
\end{longtable}
}

\subsection{Survey methodology: building and partitioning the corpus}
\label{sec:method}

\begin{figure}[t]
\centering
\resizebox{\textwidth}{!}{%
\begin{tikzpicture}[
  font=\footnotesize,
  box/.style={draw=hidden-draw, rounded corners=1pt, align=left, inner sep=5pt,
              text width=16.5em, fill=white},
  ex/.style={font=\scriptsize\itshape, text=BrickRed!85!black},
  bar/.style={draw=hidden-draw, rounded corners=1pt, inner sep=2pt, rotate=90,
              anchor=center, align=center, font=\footnotesize\bfseries, minimum width=8em},
  ar/.style={-{Stealth[length=2mm]}, darkgray, line width=0.7pt},
  node distance=4mm and 9mm,
]
\node[box, fill=secA!30] (L1)
  {\textbf{Identification (web search).} 237 candidate records from five thematic sweeps\\
   {\scriptsize\itshape\color{BrickRed!85!black} --12 cross-sweep duplicates removed}};
\node[box, fill=secC!22, below=of L1] (L2)
  {\textbf{Screening.} 225 screened against the existing corpus (bibkey, arXiv-ID)\\
   {\scriptsize\itshape\color{BrickRed!85!black} --0 records overlapping the taxonomy corpus}};
\node[box, fill=secC!22, below=of L2] (L3)
  {\textbf{Eligibility.} 225 metadata-verified: exact title, first author, year, arXiv ID\\
   {\scriptsize\itshape\color{BrickRed!85!black} --candidates failing verification dropped at harvest (count not logged)}};
\node[box, fill=secEl, below=of L3] (L4)
  {\textbf{Landscape corpus:} 225 works (Table~\ref{tab:landscape})};
\node[box, fill=secB!25, right=9mm of L1] (R1)
  {\textbf{Identification (seeding \& snowballing).} Seed systems per branch, then citation
   chasing across the six branches};
\node[box, fill=secB!12, below=of R1] (R2)
  {\textbf{Taxonomy systems:} 77 metadata-verified};
\node[box, fill=secB!12, below=of R2] (R3)
  {\textbf{Prior surveys:} 7 (venue-targeted search)};
\node[box, fill=secEl, below=7mm of L4, xshift=13em, text width=35em, align=center] (INC)
  {\textbf{Included:} 302 systems mapped (77 taxonomy $+$ 225 landscape) $+$ 7 prior surveys
   $+$ 1 industry reference $=$ \textbf{310 references}};
\draw[ar] (L1) -- (L2); \draw[ar] (L2) -- (L3); \draw[ar] (L3) -- (L4);
\draw[ar] (R1) -- (R2); \draw[ar] (R2) -- (R3);
\draw[ar] (L4.south) -- (L4.south |- INC.north);
\draw[ar] (R3.south) |- ([xshift=6em]INC.north) -- ([xshift=6em]INC.north |- INC.north);
\node[bar, fill=secA] at ($(L1.west)+(-2em,0)$) {Identification};
\node[bar, fill=secC] at ($(L2.west)!0.5!(L3.west)+(-2em,0)$) {Screening};
\node[bar, fill=secE] at ($(L4.west)+(-2em,0)$) {Included};
\end{tikzpicture}}
\caption{\textbf{Corpus construction} in the PRISMA 2020 style~\citep{prisma2020}. The left column
is the web-search stream behind the landscape corpus (Table~\ref{tab:landscape}), with recorded
counts and inline exclusions (red) at each stage; the right column is the seed-and-snowball
curation behind the weights--skills taxonomy corpus (Figure~\ref{fig:taxonomy_main}). Both streams feed the
included set. Candidates whose metadata could not be verified were discarded at harvest time; their
count was not logged, which we state rather than estimate (\S\ref{sec:method}).}
\Description{A PRISMA 2020 style flow diagram with two identification columns. The left column
(structured web search) shows 237 candidate records, minus 12 duplicates, screened to 225 with zero
taxonomy overlap, and 225 metadata-verified records forming the landscape corpus; exclusions are
noted inline, including a note that candidates failing verification were dropped with their count
not logged. The right column (seed-and-snowball curation) yields 77 taxonomy systems and 7 prior
surveys. Both columns feed an included box of 302 systems and 310 total references. Side bars mark
the Identification, Screening, and Included stages.}
\label{fig:prisma}
\end{figure}
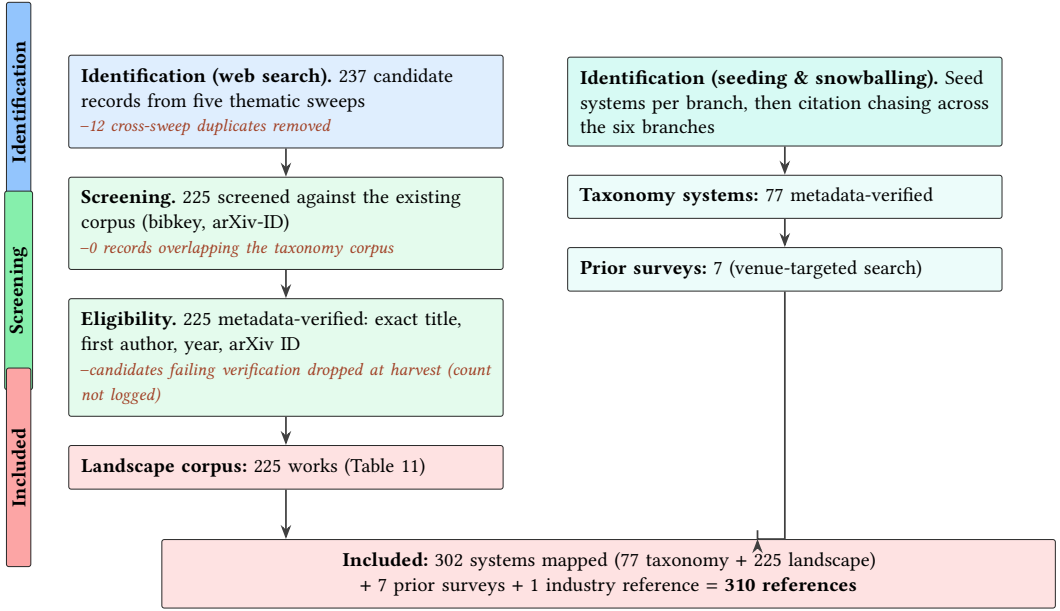

Figure~\ref{fig:prisma} summarises the corpus construction in the PRISMA~2020
style~\citep{prisma2020}. The corpus was assembled in two passes over a January~2016 to
July~2026 window. First, the 77 taxonomy systems were curated by \emph{seeding and
snowballing}: for each of the six branches we started from its landmark systems and chased
citations forward and backward until new candidates stopped appearing. Second, the 225-work
landscape corpus (Table~\ref{tab:landscape}) was gathered by five \emph{structured web-search
sweeps}, one per area cluster: (i) generalist and vision-language-action policies, imitation
and diffusion policies; (ii) language-model planning, task-and-motion planning, reward and data
synthesis; (iii) skill discovery, hierarchical reinforcement learning, and world models;
(iv) manipulation, dexterity, sim-to-real, locomotion, and benchmarks; and (v) representation
learning, offline reinforcement learning, navigation, and tactile sensing. Each sweep queried
the open web with area-specific search strings of the form ``\emph{<system or family name>
robot learning arXiv}'' and ``\emph{<area> survey/benchmark/dataset}'' over arXiv, the ACM
Digital Library, IEEE Xplore, Google Scholar, and venue proceedings pages (CoRL, RSS, ICRA,
IROS, NeurIPS, ICML, ICLR, CVPR).

\emph{Inclusion} required a work to (a) concern how an embodied agent's skill or policy is
authored, learned, transferred, evaluated, or distributed, and (b) carry verifiable metadata:
exact title, first author, year, and an arXiv identifier or DOI, each confirmed against the
source. \emph{Exclusion} removed pure perception, navigation, and locomotion work with no
bearing on skill authoring (retained only in the landscape appendix where it anchors an area),
works whose metadata could not be verified, and duplicates, detected by both bibliographic key
and arXiv identifier. Every reference in this survey passed the verification step; candidates
that failed it were discarded at harvest time, and we report that their count was not logged
rather than estimate it. Included systems were then mapped onto the taxonomy of
Figure~\ref{fig:taxonomy_main} by the mechanism definitions of Table~\ref{tab:defs}.

\paragraph{Placement: why a system enters the taxonomy or the landscape.}
Corpus membership and placement \emph{within} the corpus are two separate decisions, and it is the
second that produces the 77/225 split. A verified work is \emph{taxonomised}, that is, promoted to
the 77 systems of Figure~\ref{fig:taxonomy_main} and scored on all six axes of
Tables~\ref{tab:compare_main}--\ref{tab:compare_34}, only when it passes three tests.
(i)~\textbf{Axis-placeable}: it takes an unambiguous position on the weights-versus-skills question
and, for skill-authoring systems, on the self-improvement ladder of \S\ref{sec:code}, so that its
branch is determined rather than argued. (ii)~\textbf{Distinct}: it contributes a mechanism, rung,
or branch exemplar not already carried by a system in the map, rather than a near-variant, ablation,
or reimplementation of one. (iii)~\textbf{Fully characterisable}: its primary source reports enough
of what it \emph{ships}, how it learns, where it is evaluated, and whether it self-improves to fill
every comparison column without inference. A verified work that fails any one test is kept rather
than dropped: it stays in the landscape appendix (Table~\ref{tab:landscape}) as coverage evidence,
which is where precursors, same-point variants, and the supporting datasets, benchmarks, and
simulators live. Two further rules keep the map legible on a single page. First, when several
systems make the same taxonomic point, we taxonomise the earliest or most load-bearing landmark and
catalogue the rest in the landscape. Second, pure perception, navigation, or locomotion work is
taxonomised only when it directly authors or improves a manipulation skill, and otherwise anchors
its area in the landscape alone. The 77/225 split is therefore a statement about analytical
\emph{role}, sharp exemplar versus representative breadth, not about a work's quality or importance.

\subsection{The corpus at a glance}
\label{sec:corpus_gallery}

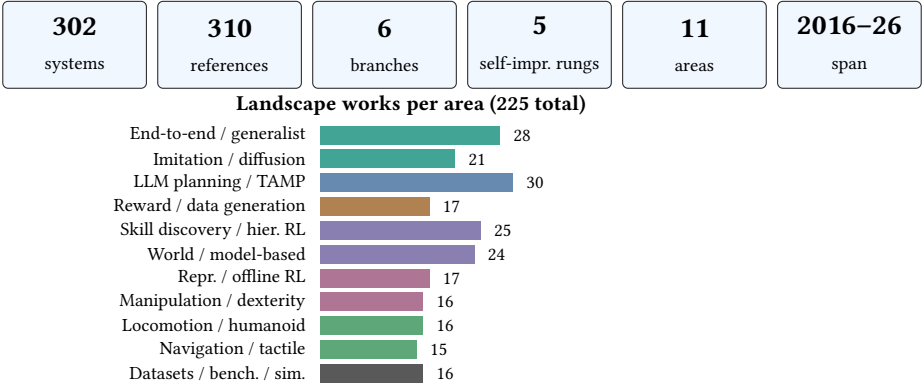
\begin{figure}[tp]
\centering
\begin{tikzpicture}
\foreach [count=\i] \num/\lab in {302/systems, 310/references, 6/branches,
                                  5/{self-impr.\ rungs}, 11/areas, {2016--26}/span}{
  \node[draw=hidden-draw, rounded corners=2pt, fill=secA!14, minimum width=1.9cm,
        minimum height=1.15cm, align=center] at ({\i*2.05-0.4},4.55)
        {{\large\bfseries\num}\\[1pt]{\scriptsize\lab}};
}
\foreach [count=\i] \l/\v/\c in {{End-to-end / generalist}/28/cat-vla,
   {Imitation / diffusion}/21/cat-vla, {LLM planning / TAMP}/30/cat-code,
   {Reward / data generation}/17/cat-rew, {Skill discovery / hier.\ RL}/25/cat-skill,
   {World / model-based}/24/cat-skill, {Repr.\ / offline RL}/17/cat-trans,
   {Manipulation / dexterity}/16/cat-trans, {Locomotion / humanoid}/16/cat-bench,
   {Navigation / tactile}/15/cat-bench, {Datasets / bench.\ / sim.}/16/gray}{
  \fill[\c!70!black] (4.9,{3.55-\i*0.315}) rectangle ({4.9+\v*0.085},{3.55-\i*0.315+0.24});
  \node[left=2pt,font=\scriptsize] at (4.9,{3.55-\i*0.315+0.12}) {\l};
  \node[right=1.5pt,font=\scriptsize] at ({4.9+\v*0.085},{3.55-\i*0.315+0.12}) {\v};
}
\node[font=\footnotesize\bfseries] at (6.1,3.75) {Landscape works per area (225 total)};
\end{tikzpicture}
\caption{\textbf{The surveyed corpus at a glance.} The survey covers 302 systems (77 placed in the
taxonomy of Figure~\ref{fig:taxonomy_main} plus 225 landscape works of Table~\ref{tab:landscape} in
Appendix~\ref{sec:landscape})
across 6 technique branches, with the code-as-policy branch resolved into 5 self-improvement rungs,
spanning 2016--2026 and totalling 310 references. The bars give the number of landscape works in
each of the 11 areas, taken directly from the area tallies of Table~\ref{tab:landscape}. The teal
and blue bars mark the survey's two poles (Figure~\ref{fig:corpus_collage}): the weights side
(end-to-end and imitation-learned policies) and the skills side (code-as-policy planning); the
remaining branches sit off that axis.}
\Description{An overview figure. Six stat callouts read: 302 systems, 310 references, 6 branches, 5
self-improvement rungs, 11 areas, span 2016 to 2026. Below, a horizontal bar chart gives landscape
works per area: End-to-end/generalist 28, Imitation/diffusion 21, LLM planning/TAMP 30,
Reward/data generation 17, Skill discovery/hierarchical RL 25, World/model-based 24,
Representation/offline RL 17, Manipulation/dexterity 16, Locomotion/humanoid 16, Navigation/tactile
15, Datasets/benchmarks/simulators 16.}
\label{fig:corpus_overview}
\end{figure}

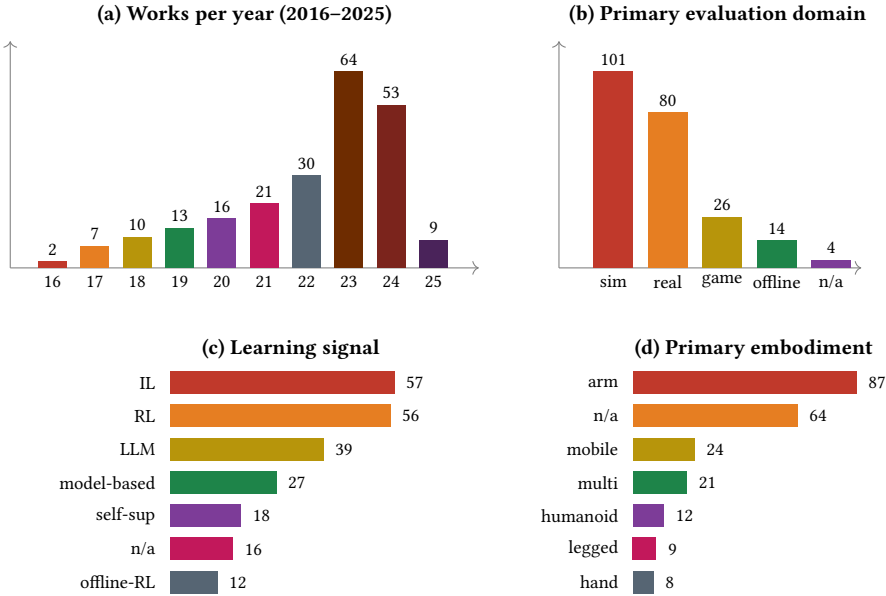
\begin{figure*}[tp]
\centering
\definecolor{pal1}{HTML}{C0392B}\definecolor{pal2}{HTML}{E67E22}\definecolor{pal3}{HTML}{B7950B}%
\definecolor{pal4}{HTML}{1E8449}\definecolor{pal5}{HTML}{7D3C98}\definecolor{pal6}{HTML}{C2185B}%
\definecolor{pal7}{HTML}{566573}\definecolor{pal8}{HTML}{6E2C00}\definecolor{pal9}{HTML}{7B241C}%
\definecolor{pal10}{HTML}{4A235A}%
\begin{tabular}{@{}c@{\hspace{7mm}}c@{}}
\begin{tikzpicture}[baseline]
  \draw[->,gray] (0,0)--(0,3.0); \draw[->,gray] (0,0)--(6.2,0);
  \foreach [count=\i] \l/\v in {16/2,17/7,18/10,19/13,20/16,21/21,22/30,23/64,24/53,25/9}{
    \fill[pal\i] ({\i*0.56-0.19},0) rectangle ({\i*0.56+0.19},{\v*0.0406});
    \node[font=\scriptsize] at ({\i*0.56},-0.18) {\l};
    \node[font=\scriptsize,above=-1pt] at ({\i*0.56},{\v*0.0406}) {\v};
  }
  \node[font=\footnotesize\bfseries] at (3.1,3.35) {(a) Works per year (2016--2025)};
\end{tikzpicture}
&
\begin{tikzpicture}[baseline]
  \draw[->,gray] (0,0)--(0,3.0); \draw[->,gray] (0,0)--(4.0,0);
  \foreach [count=\i] \l/\v/\c in {sim/101/cat-vla, real/80/cat-code, game/26/cat-rew,
                                    offline/14/cat-skill, {n/a}/4/gray}{
    \fill[pal\i] ({\i*0.72-0.26},0) rectangle ({\i*0.72+0.26},{\v*0.0257});
    \node[font=\scriptsize] at ({\i*0.72},-0.18) {\l};
    \node[font=\scriptsize,above=-1pt] at ({\i*0.72},{\v*0.0257}) {\v};
  }
  \node[font=\footnotesize\bfseries] at (2.1,3.35) {(b) Primary evaluation domain};
\end{tikzpicture}
\\[7mm]
\begin{tikzpicture}[baseline]
  \foreach [count=\i] \l/\v in {IL/57, RL/56, LLM/39, {model-based}/27, {self-sup}/18,
                                {n/a}/16, {offline-RL}/12}{
    \fill[pal\i] (0,{-\i*0.44+3.4}) rectangle ({\v*0.052},{-\i*0.44+3.7});
    \node[font=\scriptsize,left=2pt] at (0,{-\i*0.44+3.55}) {\l};
    \node[font=\scriptsize,right=1pt] at ({\v*0.052},{-\i*0.44+3.55}) {\v};
  }
  \node[font=\footnotesize\bfseries] at (1.6,3.55) {(c) Learning signal};
\end{tikzpicture}
&
\begin{tikzpicture}[baseline]
  \foreach [count=\i] \l/\v in {arm/87, {n/a}/64, mobile/24, multi/21, humanoid/12,
                                legged/9, hand/8}{
    \fill[pal\i] (0,{-\i*0.44+3.4}) rectangle ({\v*0.034},{-\i*0.44+3.7});
    \node[font=\scriptsize,left=2pt] at (0,{-\i*0.44+3.55}) {\l};
    \node[font=\scriptsize,right=1pt] at ({\v*0.034},{-\i*0.44+3.55}) {\v};
  }
  \node[font=\footnotesize\bfseries] at (1.6,3.55) {(d) Primary embodiment};
\end{tikzpicture}
\end{tabular}
\caption{\textbf{Profile of the 225 landscape works} (Table~\ref{tab:landscape},
Appendix~\ref{sec:landscape}), counted directly from the taxonomy tags. (a)~publication year, showing the surge in 2023--2024; (b)~primary
evaluation domain (simulation dominates, with real-robot a close second); (c)~primary learning
signal; (d)~primary embodiment (``n/a'' marks datasets, benchmarks, and simulators, which have
none). The four profiles cut across both poles of the weights--skills axis
(Figure~\ref{fig:corpus_collage}), so their bars use a categorical palette that avoids the reserved pole colours (teal and blue).
Counts are the exact column tallies of Table~\ref{tab:landscape}; they characterise the
surveyed corpus, not the field's total output.}
\Description{Four bar charts summarising the 225 landscape works. (a) Works per year rise from 2 in
2016 to a peak of 64 in 2023 and 53 in 2024. (b) Primary evaluation domain: 101 simulation, 80
real, 26 game, 14 offline, 4 not-applicable. (c) Primary learning signal: 57 imitation learning,
56 reinforcement learning, 39 language-model, 27 model-based, 18 self-supervised, 16
not-applicable, 12 offline reinforcement learning. (d) Primary embodiment: 87 arm, 64
not-applicable, 24 mobile, 21 multiple, 12 humanoid, 9 legged, 8 hand.}
\label{fig:corpus_dist}
\end{figure*}

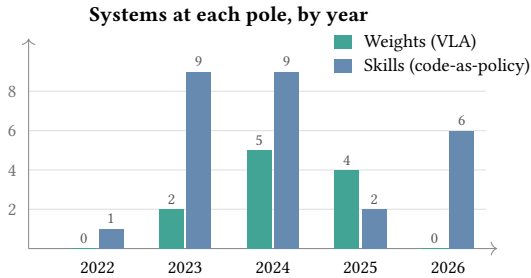
\begin{figure}[t]
\centering
\begin{tikzpicture}
  \foreach \i in {2,4,6,8}{\node[font=\scriptsize,left=1pt,text=black!55] at (0.45,{\i*0.26}) {\i};
    \draw[black!12] (0.45,{\i*0.26})--(6.5,{\i*0.26});}
  \draw[->,black!45] (0.45,0)--(0.45,2.78); \draw[->,black!45] (0.45,0)--(6.65,0);
  \foreach [count=\g] \yr/\w/\s in {2022/0/1,2023/2/9,2024/5/9,2025/4/2,2026/0/6}{
    \pgfmathsetmacro\x{\g*1.16+0.2}
    \fill[cat-vla!70!black]  ({\x-0.34},0) rectangle ({\x-0.02},{\w*0.26});
    \fill[cat-code!70!black] ({\x+0.02},0) rectangle ({\x+0.34},{\s*0.26});
    \node[font=\tiny,text=black!65,above=-1.5pt] at ({\x-0.18},{\w*0.26}) {\w};
    \node[font=\tiny,text=black!65,above=-1.5pt] at ({\x+0.18},{\s*0.26}) {\s};
    \node[font=\scriptsize] at (\x,-0.25) {\yr};}
  \fill[cat-vla!70!black]  (4.5,2.64) rectangle (4.73,2.82);
  \node[font=\scriptsize,right=1pt] at (4.73,2.73) {Weights (VLA)};
  \fill[cat-code!70!black] (4.5,2.32) rectangle (4.73,2.50);
  \node[font=\scriptsize,right=1pt] at (4.73,2.41) {Skills (code-as-policy)};
  \node[font=\footnotesize\bfseries] at (3.1,3.08) {Systems at each pole, by year};
\end{tikzpicture}
\caption{\textbf{The two poles over time.} Year of first release of the taxonomy systems this survey
places at each pole of Figure~\ref{fig:corpus_collage}: the \textbf{11} end-to-end/VLA systems that
ship \emph{weights} (\S\ref{sec:vla}; teal) and the \textbf{27} code-as-policy systems that ship
\emph{skills} (\S\ref{sec:code}; blue). Both paradigms are recent, and code-as-policy is the larger,
faster-growing pole under this survey's deep-dive; the counts reflect that coverage, not the field's
total output (2026 partial, through July). Years are from the bibliography; pole membership is the
branch assignment of Figure~\ref{fig:taxonomy_main}.}
\Description{A grouped bar chart of systems at each pole by year, 2022 to 2026. Weights (VLA), teal:
0, 2, 5, 4, 0. Skills (code-as-policy), blue: 1, 9, 9, 2, 6. Code-as-policy is the larger, faster
growing pole.}
\label{fig:pole_trend}
\end{figure}

Figure~\ref{fig:corpus_overview} summarises the whole corpus at a glance, and
Figure~\ref{fig:corpus_dist} profiles the 225 landscape works (Table~\ref{tab:landscape},
Appendix~\ref{sec:landscape}) by publication year, evaluation domain, learning signal, and
embodiment; every count is the exact tally of that table. Figure~\ref{fig:pole_trend} then charts the
two poles of Figure~\ref{fig:corpus_collage} by year of release, showing both paradigms emerging only
recently, with code-as-policy the larger and faster-growing pole.

Figures~\ref{fig:robots_canvas} and~\ref{fig:plots_canvas} then give a visual overview of the surveyed
systems: the real and simulated robots they run on (Figure~\ref{fig:robots_canvas}) and the empirical
results they report (Figure~\ref{fig:plots_canvas}); the two poles of the opening thesis figure
(Figure~\ref{fig:corpus_collage}) recur here across every embodiment. Every panel is a live hyperlink
to its source paper and is cropped from a specific figure of it; Table~\ref{tab:canvas_provenance}
records the source paper and the figure number for each, so that no image is used without a
traceable, clickable origin. The figures are vector graphics with the photographs embedded at full
resolution, so they can be zoomed without pixelation.

\input{figures/fig_robots_canvas}
\input{figures/fig_plots_canvas}

\section{The Technique Families}
\label{sec:techniques}

We now examine each of the six branches of the taxonomy (Figure~\ref{fig:taxonomy_main}) in
turn. We begin with the code-as-policy camp (\S\ref{sec:code}), the focus of this survey and
the only family we organise by \emph{degree of self-improvement}, and then survey the
weights-based (\S\ref{sec:vla}), reward-synthesis (\S\ref{sec:reward}), skill-library
(\S\ref{sec:skill}), transfer (\S\ref{sec:transfer}), and benchmark (\S\ref{sec:bench})
families that surround it.

\subsection{Code-as-Policy: Robots that Write their Own Skills}
\label{sec:code}

The organising contribution of this survey is to arrange code-as-policy methods not by
application or robot embodiment, but by \emph{how much of their own competence a system
produces at run time}. Concretely, we ask five questions in sequence: does the agent (i)
write control code at all, (ii) repair that code from execution feedback within a task,
(iii) remember validated code across tasks, (iv) search a population of programs rather
than following a single trajectory of repairs, and (v) close all of these into one
open-ended loop? Each ``yes'' is a rung on the ladder shown in the \S3.1 branch of
Figure~\ref{fig:taxonomy_main}, and
the population thins sharply toward the top: of the thirteen zero-shot systems, only a
handful ever reach the combined feedback, memory, and search regime. The purpose of this
arrangement is to name the progression and to make explicit how sparsely the top rung is
populated. We use the operational definitions of Table~\ref{tab:defs} throughout: a system
exhibits feedback, memory, or search only under the conditions stated there.

\begin{table*}[t]
\centering
\footnotesize
\renewcommand{\arraystretch}{1.3}
\begin{tabular}{@{}l p{0.42\textwidth} p{0.30\textwidth}@{}}
\toprule
\textbf{Mechanism} & \textbf{Counts (a system has it if and only if\ldots)} & \textbf{Does not count} \\
\midrule
\textbf{Feedback (F)} &
an execution-grounded signal is read and used to revise the code \emph{within} a task: a
failure detector firing, a sensory or textual summary of what went wrong, or a verifier or
reward obtained by running the program (a simulated reward qualifies when the agent uses it to
revise) &
a fresh human instruction; a fixed pre-trained value function that is never queried at run
time; a natural-language critique that is not grounded in execution \\
\textbf{Memory (M)} &
content written \emph{at run time} is stored across task boundaries and retrieved for later
tasks: a persistent code-skill library, or a retrievable episodic-text memory of past
solutions &
the model's frozen pre-trained weights; a within-task scratchpad discarded at episode end; a
fixed API library the agent cannot extend \\
\textbf{Search (S)} &
more than one candidate program is maintained, scored by execution, and selected or mutated
among: sampled-program selection, evolutionary mutation, or beam search &
a single sequential chain of repairs on one candidate (that is Feedback); one-shot sampling
with no selection; repeated prompting whose candidates are never compared \\
\bottomrule
\end{tabular}
\caption{Operational definitions of the three mechanisms used to place a system on the
code-as-policy ladder (\S\ref{sec:code}). These resolve the ambiguous boundary cases: a
failure detector or a grounded critique counts as Feedback, trained weights do not count as
Memory, and repeated prompting counts as Search only when its candidates are compared and
selected.}
\label{tab:defs}
\end{table*}

\begin{figure}[t]
\centering
\begin{tikzpicture}[
 node distance=6mm and 9mm,
 box/.style={draw=hidden-draw, rounded corners, fill=secAl, align=center,
 font=\footnotesize, minimum height=8mm, text width=15mm, inner sep=3pt},
 core/.style={box, fill=secA},
 io/.style={draw=none, font=\footnotesize\itshape},
 ar/.style={-{Stealth[length=2mm]}, darkgray, line width=0.7pt},
]
 \node[io] (task) {task};
 \node[box, fill=secA, right=of task] (actor) {actor\\ agent};
 \node[box, fill=secB, right=of actor] (eng) {exec.\\ engine (F)};
 \node[box, fill=secC, right=of eng] (mem) {skill\\ memory (M)};
 \node[box, fill=secD, right=of mem] (srch) {evol.\\ search (S)};
 \node[io, right=of srch] (out) {validated skill};

 \draw[ar] (task) -- (actor);
 \draw[ar] (actor) -- (eng);
 \draw[ar] (eng) -- (mem);
 \draw[ar] (mem) -- (srch);
 \draw[ar] (srch) -- (out);
 \draw[ar] (eng.south) to[out=-90,in=-90] node[below,font=\scriptsize]{traces} (actor.south);
 \draw[ar] (mem.north) to[out=90,in=90] node[above,font=\scriptsize]{skills $\rightarrow$ future tasks} (actor.north);
\end{tikzpicture}
\caption{The full self-improving loop (\S3.1.5): feedback (F) $+$ memory (M) $+$ search (S)
combine into one open-ended loop (occupied by ASPIRE, ENPIRE, and RoboClaw). VLAs (\S3.2) have no such loop.}
\Description{A left-to-right pipeline of five boxes: task, actor agent, execution engine (feedback),
skill memory, and evolutionary search, producing a validated skill; two return arrows feed execution
traces back to the actor and store learned skills for future tasks, forming a closed self-improving loop.}
\label{fig:architectures}
\end{figure}
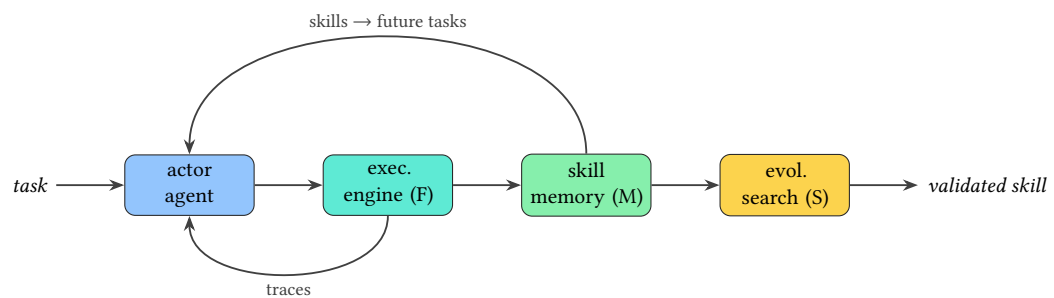

\subsubsection{Zero-shot synthesis}
\label{ssec:zeroshot}
\emph{Membership condition:} a system is on this rung if and only if it emits executable control
code but closes no execution loop back to code generation, so feedback, memory, and search are
all absent. The foundational move is to treat the language model as a \emph{program synthesiser} over
a fixed library of perception and control primitives. \emph{Code-as-Policies}
\citep{codeaspolicies} showed that an LLM prompted with an API and a natural-language
instruction can emit executable Python that composes those primitives, recursively
defining undefined functions and parameterising control with arithmetic and feedback
logic. \emph{ProgPrompt} \citep{progprompt} situates the generated program in the scene
by exposing available actions and objects as importable symbols, while \emph{VoxPoser}
\citep{voxposer} has the model write code that \emph{composes 3D value maps} for a motion
planner rather than calling fixed skills, loosening the dependence on a hand-authored API.
Subsequent work broadened the input modality and task horizon; \emph{RoboCodeX}
\citep{robocodex} conditions on multimodal observations, \emph{Instruct2Act}
\citep{instruct2act} maps instructions to perception--action code, and video- and
demonstration-conditioned variants such as \emph{RoboPro} \citep{robopro},
\emph{Demo2Code} \citep{demo2code} and \emph{Text2Motion} \citep{text2motion} recover
programs from richer context. \emph{Statler} \citep{statler} maintains an explicit world
state across steps, and \emph{TidyBot} \citep{tidybot} and \emph{ChatGPT for Robotics}
\citep{chatgptrobotics} demonstrate personalisation and prompt design for real hardware.
What unites this rung, and limits it, is that the program is written \emph{once}: there is
no channel from execution back to the synthesiser, so a plan that fails at run time simply
fails.

\subsubsection{Closed-loop self-repair}
\label{ssec:selfrepair}
\emph{Membership condition:} feedback is present but memory and search are absent; the agent
revises code within a task from an execution-grounded signal yet retains nothing across tasks
and never compares a population of candidates. The second rung adds a within-task feedback loop. \emph{Inner Monologue}
\citep{innermonologue} feeds success detectors, scene descriptions and human feedback back
into the model as language, letting it replan when a step fails. \emph{DoReMi}
\citep{doremi} detects misalignments between plan and execution and triggers recovery, and
\emph{REFLECT} \citep{reflect} builds a hierarchical, multimodal summary of past
interaction that an LLM queries to explain a failure and propose a correction. More recent
systems push the monitor into the perception loop; \emph{Code-as-Monitor}
\citep{codeasmonitor} compiles spatio-temporal constraints into vision-language checks, and
\emph{AHA} \citep{aha} trains a model to reason explicitly about failure modes. The novelty
of this rung is the closed loop, but it is a \emph{short} loop: corrections are discarded
at task boundaries, so nothing accumulates.

\subsubsection{Skill-library accumulation}
\label{ssec:skilllib}
\emph{Membership condition:} memory is present; validated code is written to a cross-task store
and retrieved later, independently of whether within-task feedback is used. The third rung makes the loop persistent by writing validated code into a growing library
that later tasks retrieve and compose. Because this ``memory'' axis is a full technique
family in its own right, we catalogue its systems once, in \S\ref{sec:skill}
(\S3.4.3); \emph{Voyager} \citep{voyager} is the canonical example, curating an
ever-growing library of verified skill programs with when-to-apply guards, while related
methods distil language corrections into retrievable knowledge for future tasks~\citep{droc}. The key
distinction from \S3.1.2 is temporal scope: competence compounds \emph{across} tasks rather
than being rebuilt each time.

\subsubsection{Evolutionary program search}
\label{ssec:search}
\emph{Membership condition:} search is present; the agent maintains and selects among more than
one candidate program rather than repairing a single one. The fourth rung replaces a single chain of repairs with population-based search over
programs. Rather than iteratively debugging one candidate, the agent maintains several,
scores them by execution, and mutates the best. \emph{CaP-X} \citep{capx} provides an
interactive gym and benchmark for exactly this style of coding agent, and \emph{RoboEvolve}
\citep{roboevolve} couples a planner and a learned simulator into a co-evolutionary loop
that mines near-miss failures to stabilise search. The novelty is the shift from
\emph{repair} to \emph{search}: exploration is explicit and parallel, which trades compute
for a better chance of escaping local failures.

\subsubsection{The full self-improving loop}
\label{ssec:fullloop}
\emph{Membership condition:} all three mechanisms are present in one closed loop, so the rung
is the conjunction feedback \emph{and} memory \emph{and} search (Figure~\ref{fig:architectures}).
Only a few very recent systems satisfy it. \emph{ASPIRE} \citep{aspire} pairs a robot execution
engine that emits per-primitive multimodal traces with a skill library of validated code and a
population search over candidate programs; \emph{ENPIRE} \citep{enpire} closes an analogous loop
on physical hardware through automatic reset, parallel rollouts, and log-driven revision; and
\emph{RoboClaw} \citep{roboclaw} unifies data collection, policy learning, and execution under a
single controller with self-resetting loops. That this cell is so sparsely populated, rather
than any particular system within it, is the structural observation the survey is organised to
make.

Table~\ref{tab:compare_31} makes the funnel explicit, tabulating each system's use of
feedback (F), memory (M), and search (S).
{\footnotesize
\begin{longtable}{@{}p{0.36\textwidth}c ccc c@{}}
\caption{The 30 systems on the code-as-policy self-improvement ladder: the 27 of the \S3.1 branch of
Figure~\ref{fig:taxonomy_main} plus the three \S3.4.3 skill-library systems (Voyager, LRLL, Uni-Skill)
that also occupy its memory rung, on the three
self-improvement mechanisms, \textbf{F}eedback, \textbf{M}emory, \textbf{S}earch, which are
determined by the rung. \yy\ present, \nn\ absent. \textbf{Self-impr.}: \yy\ full loop,
\pp\ partial, \nn\ none.}\label{tab:compare_31}\\
\toprule
\textbf{System} & \textbf{Year} & \textbf{F} & \textbf{M} & \textbf{S} & \textbf{Self-impr.}\\
\midrule
\endfirsthead
\multicolumn{6}{@{}l}{\footnotesize\emph{Table~\ref{tab:compare_31} (continued)}}\\
\toprule
\textbf{System} & \textbf{Year} & \textbf{F} & \textbf{M} & \textbf{S} & \textbf{Self-impr.}\\
\midrule
\endhead
\bottomrule
\endlastfoot
\multicolumn{6}{@{}l}{\textbf{\S3.1.1 Zero-shot synthesis} \hfill F\,\nn\ \ M\,\nn\ \ S\,\nn}\\
\textbf{Code-as-Policies}~\cite{codeaspolicies} & 2023 & \nn & \nn & \nn & \nn\\
\textbf{ProgPrompt}~\cite{progprompt} & 2023 & \nn & \nn & \nn & \nn\\
\textbf{VoxPoser}~\cite{voxposer} & 2023 & \nn & \nn & \nn & \nn\\
\textbf{Instruct2Act}~\cite{instruct2act} & 2023 & \nn & \nn & \nn & \nn\\
\textbf{ChatGPT for Robotics}~\cite{chatgptrobotics} & 2023 & \nn & \nn & \nn & \nn\\
\textbf{TidyBot}~\cite{tidybot} & 2023 & \nn & \nn & \nn & \nn\\
\textbf{RoboCodeX}~\cite{robocodex} & 2024 & \nn & \nn & \nn & \nn\\
\textbf{RoboScript}~\cite{roboscript} & 2024 & \nn & \nn & \nn & \nn\\
\textbf{Text2Motion}~\cite{text2motion} & 2023 & \nn & \nn & \nn & \nn\\
\textbf{Demo2Code}~\cite{demo2code} & 2023 & \nn & \nn & \nn & \nn\\
\textbf{Statler}~\cite{statler} & 2024 & \nn & \nn & \nn & \nn\\
\textbf{Prompt2Walk}~\cite{prompt2walk} & 2024 & \nn & \nn & \nn & \nn\\
\textbf{RoboPro}~\cite{robopro} & 2025 & \nn & \nn & \nn & \nn\\
\midrule
\multicolumn{6}{@{}l}{\textbf{\S3.1.2 Closed-loop self-repair} \hfill F\,\yy\ \ M\,\nn\ \ S\,\nn}\\
\textbf{Inner Monologue}~\cite{innermonologue} & 2022 & \yy & \nn & \nn & \nn\\
\textbf{DoReMi}~\cite{doremi} & 2024 & \yy & \nn & \nn & \nn\\
\textbf{REFLECT}~\cite{reflect} & 2023 & \yy & \nn & \nn & \nn\\
\textbf{Code-as-Monitor}~\cite{codeasmonitor} & 2025 & \yy & \nn & \nn & \nn\\
\textbf{AHA}~\cite{aha} & 2024 & \yy & \nn & \nn & \nn\\
\textbf{Introspective Planning}~\cite{introspective} & 2024 & \yy & \nn & \nn & \nn\\
\midrule
\multicolumn{6}{@{}l}{\textbf{\S3.1.3 Skill-library accumulation} \hfill F\,\nn\ \ M\,\yy\ \ S\,\nn}\\
\textbf{Voyager}~\cite{voyager} & 2023 & \nn & \yy & \nn & \pp\\
\textbf{LRLL}~\cite{lrll} & 2024 & \nn & \yy & \nn & \pp\\
\textbf{RoboCoder}~\cite{robocoder} & 2024 & \nn & \yy & \nn & \pp\\
\textbf{Uni-Skill}~\cite{uniskill} & 2026 & \nn & \yy & \nn & \pp\\
\textbf{DROC}~\cite{droc} & 2024 & \yy & \yy & \nn & \pp\\
\midrule
\multicolumn{6}{@{}l}{\textbf{\S3.1.4 Evolutionary program search} \hfill F\,\yy\ \ M\,\nn\ \ S\,\yy}\\
\textbf{CaP-X}~\cite{capx} & 2026 & \yy & \nn & \yy & \pp\\
\textbf{RoboEvolve}~\cite{roboevolve} & 2026 & \yy & \nn & \yy & \pp\\
\textbf{Code Evolution (GEAR)}~\cite{codeevolution} & 2026 & \yy & \nn & \yy & \pp\\
\midrule
\multicolumn{6}{@{}l}{\textbf{\S3.1.5 Full self-improving loop} \hfill F\,\yy\ \ M\,\yy\ \ S\,\yy}\\
\textbf{ASPIRE}~\cite{aspire} & 2026 & \yy & \yy & \yy & \yy\\
\textbf{ENPIRE}~\cite{enpire} & 2026 & \yy & \yy & \yy & \yy\\
\textbf{RoboClaw}~\cite{roboclaw} & 2026 & \yy & \yy & \yy & \yy\\
\end{longtable}
}

\subsection{End-to-End Vision-Language-Action Models}
\label{sec:vla}

The dominant alternative to writing code is to regress actions directly from pixels and
language with a single large network, so that competence lives in \emph{frozen weights}
rather than in an inspectable program. Architecturally these vision-language-action (VLA)
models share a backbone--action-head decomposition: a vision-language backbone encodes the
scene and instruction, and a pluggable action head maps that representation to motor
commands~\citep{openvla,pi0}.

The lineage begins with \emph{RT-1}~\citep{rt1}, a transformer trained on large-scale real
robot data that discretises continuous actions into categorical bins. \emph{RT-2}~\citep{rt2}
then co-fine-tunes an Internet-pretrained vision-language model on robot trajectories,
emitting actions as text tokens and inheriting semantic generalisation from web data. Open
reproductions followed: \emph{Octo}~\citep{octo} and \emph{OpenVLA}~\citep{openvla}, the
latter a 7B model with a DINOv2/SigLIP vision stack, are trained on the pooled
Open X-Embodiment corpus (\S\ref{sec:transfer}), and \emph{RoboFlamingo}~\citep{roboflamingo}
adapts a vision-language model into an imitation policy.

A second design axis is the \emph{action representation}. Discrete token heads (RT-1/RT-2)
are simple but coarse; recent systems favour continuous heads. \emph{$\pi_0$}~\citep{pi0}
attaches a \emph{flow-matching} head to a PaLI-Gemma backbone for smooth, high-frequency
bimanual control, and \emph{$\pi_{0.5}$}~\citep{pi05} extends it toward open-world
generalisation; \emph{CogACT}~\citep{cogact} separates a cognition module from a
diffusion-transformer action module, and \emph{SpatialVLA}~\citep{spatialvla} injects 3D
spatial encodings. Flow- and diffusion-based heads trade a few extra parameters for far fewer
inference steps and better high-degree-of-freedom precision.

At the largest scale, foundation-model efforts target whole platforms:
\emph{GR00T~N1}~\citep{groot} pairs a slow System-2 vision-language reasoner with a fast
System-1 diffusion transformer that produces motor actions at high rate for humanoids, and
\emph{Gemini Robotics}~\citep{geminirobotics} builds a VLA together with an embodied-reasoning
model on the Gemini backbone. Across all of these, the defining property, and the reason we
treat them as the \emph{foil} for this survey, is that there is \emph{no loop and no code}:
competence is acquired by scaling data and parameters, generalises by interpolation rather
than search, and cannot be edited, audited, or recombined after training.

\subsection{Reward and Curriculum Synthesis}
\label{sec:reward}

A third family keeps the language model in the role of programmer but points it at a
different artefact: instead of policy code, it writes the \emph{reward}, \emph{environment},
or \emph{curriculum} that a reinforcement-learning loop then compiles into a policy. A
feedback loop exists, the model reads back training statistics and revises, but what ships
is still weights, which is why we separate this family from the self-improving code camp
(\S\ref{sec:code}).

\emph{Eureka}~\citep{eureka} is the canonical instance: given unmodified environment source
code and a task description, a coding LLM zero-shot generates an executable reward function,
then improves it by \emph{evolutionary search} guided by \emph{reward reflection}, a textual
summary of training statistics, exceeding expert human rewards on 83\% of a 29-task IsaacGym
suite. \emph{DrEureka}~\citep{dreureka} extends the idea to sim-to-real by also synthesising
domain-randomisation ranges, achieving zero-shot real-world locomotion.
\emph{Text2Reward}~\citep{text2reward} generates dense reward programs grounded in a compact
environment representation and refines them from human feedback, matching or beating
expert-written rewards on most of a seventeen-task suite, while \emph{Language to
Rewards}~\citep{l2r} uses the reward function as the interface between an LLM and a MuJoCo
model-predictive controller.

Beyond the reward itself, the same generative recipe is applied to the \emph{training
environment}. \emph{Eurekaverse}~\citep{eurekaverse} has an LLM propose a progressively harder
curriculum of environments, learning parkour skills that transfer to a real robot, and
\emph{RoboGen}~\citep{robogen} runs a propose-generate-learn cycle that fabricates tasks,
scenes, and supervision in simulation to unlock effectively unlimited training data;
\emph{Auto MC-Reward}~\citep{automcreward} automates dense rewards for open-world Minecraft.
The through-line is that language becomes a bridge from high-level intent to low-level control
\emph{via} an RL optimiser, rather than producing an executable, reusable skill directly.

\subsection{Skill Libraries: How a Repertoire is Built}
\label{sec:skill}

Where \S\ref{sec:code} asked \emph{how} a code skill improves, this family asks the prior
question of \emph{how a repertoire of skills is built at all}. The \S3.4 branch of
Figure~\ref{fig:taxonomy_main} organises these systems along one axis, skills learned as latent reinforcement-learning policies
versus skills stored as retrievable code, and only the code flavour self-improves without
gradients, which links this family back to \S\ref{sec:code}.

Before surveying these systems we note that the word ``skill'' is badly overloaded.
Table~\ref{tab:skill_senses} separates five senses that the literature routinely conflates,
from a latent policy to a market product, and records which properties each sense affords.

\begin{table*}[t]
\centering
\footnotesize
\renewcommand{\arraystretch}{1.25}
\begin{tabular}{@{}p{0.16\textwidth} p{0.22\textwidth} p{0.18\textwidth} c c c c@{}}
\toprule
\textbf{Sense of ``skill''} & \textbf{Representation} & \textbf{Examples} &
\textbf{Inspect.} & \textbf{Adapt.} & \textbf{Compos.} & \textbf{Distrib.} \\
\midrule
latent policy & latent-conditioned network $\pi(a\mid s,z)$ & DIAYN, DADS, METRA & \nn & \pp & \nn & \nn \\
option / primitive & temporally-extended sub-policy with a skill prior & SPiRL, OPAL, SkiMo & \pp & \pp & \yy & \nn \\
code & executable program over an API & Code-as-Policies, Voyager & \yy & \yy & \yy & \pp \\
robot app & packaged, one-tap deployable behaviour & UniStore motion packages & \nn & \nn & \nn & \yy \\
market product & a listed, versioned, priced commodity & UniStore listings & \nn & \nn & \pp & \yy \\
\bottomrule
\end{tabular}
\caption{The word ``skill'' spans at least five senses that the literature frequently
conflates. Columns record whether a skill is human-\textbf{inspect}able, \textbf{adapt}able
after creation, symbolically \textbf{compos}able, and \textbf{distrib}utable across robots
(\yy\ yes, \pp\ partial, \nn\ no). Only the \emph{code} sense is inspectable, adaptable, and
composable at once; only the app and market senses are readily distributable, which is the
mismatch the skill economy (\S\ref{sec:open}) must close.}
\label{tab:skill_senses}
\end{table*}

\subsubsection{Unsupervised and latent skill discovery}
The oldest lineage learns a diverse set of skills with \emph{no task reward and no code}, by
maximising an information-theoretic objective. \emph{DIAYN}~\citep{diayn} maximises the mutual
information between a latent skill code and the visited states, yielding distinguishable
behaviours; \emph{DADS}~\citep{dads} makes the discovered skills dynamics-aware and hence
usable for model-based control; and \emph{LSD}~\citep{lsd}, \emph{CIC}~\citep{cic}, and
\emph{METRA}~\citep{metra} push toward more dynamic, far-reaching, and scalable skills through
Lipschitz constraints, contrastive objectives, and metric-aware abstractions respectively.

\subsubsection{Skill-based and hierarchical reinforcement learning}
A second lineage extracts a continuous \emph{skill space} from offline data and then plans or
learns within it. \emph{SPiRL}~\citep{spirl} learns a skill embedding together with a skill
prior that accelerates downstream RL; \emph{OPAL}~\citep{opal} discovers temporally extended
primitives for offline RL; \emph{PARROT}~\citep{parrot} learns an invertible behavioural
prior; and \emph{SkiMo}~\citep{skimo} learns a skill dynamics model so that planning happens
directly in skill space.

\subsubsection{LLM and code skill libraries}
The code flavour, where the self-improving methods of \S\ref{sec:code} live, stores skills as
retrievable programs. \emph{Voyager}~\citep{voyager} curates an ever-growing library of
verified code skills in Minecraft; \emph{LOTUS}~\citep{lotus} performs continual imitation
learning through unsupervised skill discovery; \emph{LRLL}~\citep{lrll} bootstraps a composable
robot library with language models; and \emph{BOSS}~\citep{boss} and \emph{SPRINT}~\citep{sprint}
grow skill repertoires by LLM-guided bootstrapping and instruction relabelling. The most recent
entries close the loop between video and code: \emph{Uni-Skill}~\citep{uniskill} builds a
self-evolving skill repository from unstructured robot video, and \emph{SkillFlow}~\citep{skillflow}
benchmarks lifelong discovery, patching, and reuse of a skill library.

\subsubsection{Open-world LLM-agent libraries}
Finally, embodied LLM agents accumulate skills from open-ended experience.
\emph{GITM}~\citep{gitm} and \emph{JARVIS-1}~\citep{jarvis1} pair a planner with text or
multimodal memory in Minecraft; \emph{ExpeL}~\citep{expel} extracts reusable insights from a
stream of trials; \emph{Optimus-1}~\citep{optimus1} adds a hybrid multimodal memory for
long-horizon tasks; and \emph{Odyssey}~\citep{odyssey} equips an agent with an open-world
library of primitive and compositional skills. These systems foreshadow the marketplace
dynamics of \S\ref{sec:open}: a library that grows with use.

Table~\ref{tab:compare_34} contrasts the four sub-families on skill form, whether they are
gradient-trained or LLM-authored, whether they keep a persistent library, and their domain.
{\footnotesize
\begin{longtable}{@{}p{0.30\textwidth}c l c c c@{}}
\caption{The 21 skill-building systems of the \S3.4 branch of Figure~\ref{fig:taxonomy_main}, by sub-family.
\textbf{Form}: \flatent/\fspace/\fcodelib/\fagent. \textbf{Grad.}: gradient-trained (\yy) vs
LLM-authored (\nn). \textbf{Lib.}: persistent retrievable library. \textbf{Domain}:
\dsim/\dgame/\dreal/\dtext.}\label{tab:compare_34}\\
\toprule
\textbf{System} & \textbf{Year} & \textbf{Form} & \textbf{Grad.} & \textbf{Lib.} & \textbf{Domain}\\
\midrule
\endfirsthead
\multicolumn{6}{@{}l}{\footnotesize\emph{Table~\ref{tab:compare_34} (continued)}}\\
\toprule
\textbf{System} & \textbf{Year} & \textbf{Form} & \textbf{Grad.} & \textbf{Lib.} & \textbf{Domain}\\
\midrule
\endhead
\bottomrule
\endlastfoot
\multicolumn{6}{@{}l}{\textbf{\S3.4.1 Unsupervised / latent discovery (RL)}}\\
\textbf{DIAYN}~\cite{diayn} & 2019 & \flatent & \yy & \nn & \dsim\\
\textbf{DADS}~\cite{dads} & 2020 & \flatent & \yy & \nn & \dsim\\
\textbf{LSD}~\cite{lsd} & 2022 & \flatent & \yy & \nn & \dsim\\
\textbf{CIC}~\cite{cic} & 2022 & \flatent & \yy & \nn & \dsim\\
\textbf{METRA}~\cite{metra} & 2024 & \flatent & \yy & \nn & \dsim\\
\midrule
\multicolumn{6}{@{}l}{\textbf{\S3.4.2 Skill-based / hierarchical RL}}\\
\textbf{SPiRL}~\cite{spirl} & 2020 & \fspace & \yy & \nn & \dsim\\
\textbf{OPAL}~\cite{opal} & 2021 & \fspace & \yy & \nn & \dsim\\
\textbf{PARROT}~\cite{parrot} & 2021 & \fspace & \yy & \nn & \dsim\\
\textbf{SkiMo}~\cite{skimo} & 2022 & \fspace & \yy & \nn & \dsim\\
\midrule
\multicolumn{6}{@{}l}{\textbf{\S3.4.3 LLM / code skill libraries}}\\
\textbf{Voyager}~\cite{voyager} & 2023 & \fcodelib & \nn & \yy & \dgame\\
\textbf{LOTUS}~\cite{lotus} & 2024 & \fcodelib & \nn & \yy & \dsim\\
\textbf{LRLL}~\cite{lrll} & 2024 & \fcodelib & \nn & \yy & \dsim\\
\textbf{BOSS}~\cite{boss} & 2023 & \fcodelib & \nn & \yy & \dsim\\
\textbf{SPRINT}~\cite{sprint} & 2024 & \fcodelib & \nn & \yy & \dsim\\
\textbf{Uni-Skill}~\cite{uniskill} & 2026 & \fcodelib & \nn & \yy & \dreal\\
\textbf{SkillFlow}~\cite{skillflow} & 2026 & \fcodelib & \nn & \yy & \dtext\\
\midrule
\multicolumn{6}{@{}l}{\textbf{\S3.4.4 Open-world LLM-agent libraries}}\\
\textbf{GITM}~\cite{gitm} & 2023 & \fagent & \nn & \yy & \dgame\\
\textbf{JARVIS-1}~\cite{jarvis1} & 2023 & \fagent & \nn & \yy & \dgame\\
\textbf{ExpeL}~\cite{expel} & 2024 & \fagent & \nn & \yy & \dtext\\
\textbf{Optimus-1}~\cite{optimus1} & 2024 & \fagent & \nn & \yy & \dgame\\
\textbf{Odyssey}~\cite{odyssey} & 2024 & \fagent & \nn & \yy & \dgame\\
\end{longtable}
}

\subsection{Sim-to-Real and Cross-Embodiment Transfer}
\label{sec:transfer}

A skill is only useful if it moves across the simulation-to-reality gap and across robot
bodies. The enabling artefact is scale: \emph{Open X-Embodiment}~\citep{openx} pools sixty
datasets from twenty-one institutions into over a million trajectories spanning twenty-two
embodiments and hundreds of skills, and shows that RT-X models trained on the mixture exhibit
positive transfer and emergent capabilities across platforms.

Given such data, a single network can control very different bodies.
\emph{CrossFormer}~\citep{crossformer} trains one transformer on 900K trajectories across
twenty embodiments, arms, wheeled robots, quadrotors, and quadrupeds, \emph{without} manually
aligning observation or action spaces, and matches specialist policies tailored to each robot.
Other work attacks transfer at test time rather than through data:
\emph{Mirage}~\citep{mirage} achieves \emph{zero-shot} cross-embodiment transfer by
``cross-painting'', masking the target robot and inpainting the source robot at the same pose
so the policy sees a familiar body, and bridging the control gap with forward dynamics.

The most suggestive point for this survey is \emph{RoboCat}~\citep{robocat}, a goal-conditioned
multi-embodiment agent that adapts to a new task from as few as a hundred demonstrations and
then \emph{generates its own data} for the next training round, forming a rudimentary
self-improvement loop of the kind \S\ref{sec:code} makes explicit. This is also the marketplace
question of \S\ref{sec:open}: a skill advertised for the G1, H1, B2, and Go2
platforms must survive exactly the cross-embodiment gap these methods study.

\subsection{Benchmarks and Evaluation}
\label{sec:bench}

Progress in the families above is measured on a shared set of simulated benchmarks, most of
which report a single-number success rate. \emph{LIBERO}~\citep{libero} targets lifelong
manipulation and knowledge transfer across task suites; \emph{Meta-World}~\citep{metaworld}
offers fifty manipulation tasks for multi-task and meta-reinforcement learning;
\emph{RLBench}~\citep{rlbench} provides a hundred vision-based tasks with demonstrations; and
\emph{ManiSkill2}~\citep{maniskill2} adds GPU-parallel simulation over twenty task families.
\emph{Robosuite}~\citep{robosuite} is the modular MuJoCo framework underlying many of these,
and \emph{CALVIN}~\citep{calvin} evaluates long-horizon, language-conditioned control by the
average number of sub-tasks completed in a chain. At the high-realism end,
\emph{BEHAVIOR-1K}~\citep{behavior1k} specifies a thousand everyday household activities in a
photorealistic simulator, scoring both success and efficiency.

Table~\ref{tab:matrix} contrasts one exemplar per camp on the axes that separate them. The
recurring gap, and a direct motivation for the open problems of \S\ref{sec:open}, is that
these suites measure \emph{one-shot competence}: they report whether a policy succeeds, not
whether it \emph{improves with experience}. No standard benchmark yet plots held-out success as
a function of accumulated interaction, which is exactly the quantity the self-improvement ladder
of \S\ref{sec:code} is designed to raise.

\begin{table*}[t]
\centering
\footnotesize
\renewcommand{\arraystretch}{1.25}
\begin{tabular}{@{}p{0.25\textwidth}cccc@{}}
\toprule
\textbf{Dimension}
 & \textcolor{cat-code!50!black}{\textbf{Full loop}}~\cite{aspire,enpire,roboclaw}
 & \textbf{$\pi_0$}~\cite{pi0}
 & \textbf{Code-as-Pol.}~\cite{codeaspolicies}
 & \textbf{Eureka}~\cite{eureka} \\
 & {\footnotesize\itshape code+skills} & {\footnotesize\itshape VLA weights}
 & {\footnotesize\itshape code, no mem.} & {\footnotesize\itshape reward} \\
\midrule
What it outputs & skill code+lib & action chunks & policy code & reward code \\
How it improves & F+M+S loop & gradient pretrain & \dm\,one-shot & RL+reflection \\
Gradient training? & \nn & \yy & \nn & \yy \\
Persistent skill library? & \yy & \nn & \nn & \nn \\
Failure feedback signal & exec.\ traces & \dm & \dm & train.\ stats \\
Search strategy & evolutionary & \dm & \dm & \dm \\
Real-robot validation & \yy & \yy & \yy & \yy \\
Continual / open-ended? & \yy & \nn & \nn & \nn \\
Beyond fixed APIs? & \yy & \yy & \nn & \dm \\
Interpretable / editable? & \yy & \nn & \yy & \yy \\
\bottomrule
\end{tabular}
\caption{Capability matrix, one representative per camp. Only the full-loop cell (\S3.1.5; e.g.\ ASPIRE,
ENPIRE, RoboClaw) combines a persistent skill library with feedback- and search-driven
self-improvement; each other camp lacks at least one axis. \yy~= yes, \nn~= no, \dm~= n/a.}
\label{tab:matrix}
\end{table*}

\subsection{Synthesis: what each family cannot do}
\label{sec:synthesis}
Table~\ref{tab:branch_limits} compares every family on six axes: data needs, task horizon,
transfer, interpretability, safety, and characteristic failure mode. Reading it column-wise
profiles a family; reading it row-wise shows that every axis has at least one weak family, and
no family is favourable on all six, which is the tension the skill economy (\S\ref{sec:open})
inherits.

\begin{table*}[t]
\centering
\scriptsize
\setlength{\tabcolsep}{3pt}
\renewcommand{\arraystretch}{1.45}
\begin{tabular}{@{}>{\raggedright\arraybackslash}p{0.125\textwidth} >{\raggedright\arraybackslash}p{0.16\textwidth} >{\raggedright\arraybackslash}p{0.16\textwidth} >{\raggedright\arraybackslash}p{0.15\textwidth} >{\raggedright\arraybackslash}p{0.15\textwidth} >{\raggedright\arraybackslash}p{0.155\textwidth}@{}}
\toprule
 & \textbf{End-to-end VLA} (\S\ref{sec:vla}) & \textbf{Code-as-policy} (\S\ref{sec:code})
 & \textbf{Reward synthesis} (\S\ref{sec:reward}) & \textbf{RL skill discovery} (\S\ref{sec:skill})
 & \textbf{Market skills} (\S\ref{sec:open}) \\
\midrule
\textbf{Data needs} &
\nn\ web-scale teleop corpora: 1M+ episodes, 22 embodiments~\citep{openx} &
\yy\ few-shot prompts over a pretrained LLM~\citep{codeaspolicies} &
\pp\ massive \emph{simulated} interaction, no demonstrations~\citep{eureka} &
\nn\ millions of environment steps per embodiment~\citep{diayn} &
\yy\ none at deployment; vendor-recorded playback~\citep{unistore2026} \\
\textbf{Task horizon} &
\pp\ short reactive chunks; long tasks by chaining~\citep{rt2} &
\yy\ long: programs compose perception and control primitives~\citep{voxposer} &
\nn\ one low-level skill per synthesised reward~\citep{eureka} &
\nn\ temporally-extended primitives, not full tasks~\citep{metra} &
\pp\ fixed multi-step routines; no branching \\
\textbf{Transfer} &
\yy\ cross-task and cross-embodiment via co-training~\citep{openx} &
\pp\ API-level portability, but bound to the perception stack &
\pp\ sim-to-real via randomisation search~\citep{dreureka} &
\pp\ latent skills seed downstream RL~\citep{spirl} &
\nn\ certified robot models only; no adaptation~\citep{unistore2026} \\
\textbf{Interpretability} &
\nn\ opaque weights; behaviour only observable &
\yy\ readable, editable, diffable programs &
\pp\ readable reward code; opaque trained policy &
\nn\ latent skill codes, symbolically uninspectable &
\nn\ sealed vendor package \\
\textbf{Safety} &
\nn\ emergent behaviour is hard to verify pre-deployment &
\yy\ inspectable before execution; constrained to vetted APIs &
\nn\ reward hacking; unintended optima~\citep{eureka} &
\nn\ unconstrained exploration unsafe on hardware &
\pp\ vendor certification, but no provenance record \\
\textbf{Failure mode} &
silent misgrounding under distribution shift &
brittle when perception mislabels the scene &
specification gaming that passes in sim only &
skill collapse into degenerate behaviours &
context mismatch with no recovery path \\
\bottomrule
\end{tabular}
\caption{\textbf{Family comparison matrix}: the five technique families on six axes (data needs,
task horizon, transfer, interpretability, safety, characteristic failure mode). Marks grade each
family on the axis (\yy\ favourable, \pp\ mixed, \nn\ weak); evidence anchors include the
Open X-Embodiment corpus~\citep{openx} for data and transfer and lifelong suites such as
LIBERO~\citep{libero} for horizon and reuse. No family is favourable on all six axes, which is
the tension the skill economy (\S\ref{sec:open}) inherits.}
\label{tab:branch_limits}
\end{table*}

\section{The Skill Economy: Open Problems}
\label{sec:open}

The arrival of a commercial marketplace for robot skills (Unitree's UniStore ships
one-tap, cross-model motion packages~\citep{unistore2026}) turns several research
questions from hypothetical into pressing. We frame them as the open agenda of this survey.

\paragraph{The static-to-adaptive gap.}
Every skill shipped today is, in effect, replayed: a motion package or a fixed program is
installed and executed without perceiving whether \emph{this} kitchen, gripper, or object
differs from the one it was authored for. The techniques of \S\ref{sec:code}, within-task
repair, persistent memory, and search, are precisely what would let a downloaded skill
\emph{adapt} on the target robot. Closing this gap is the difference between a store of
animations and a store of capabilities, and it is the direct application of the \S3.1.5
loop to a distribution setting where the base skill is given rather than discovered.

\paragraph{Cross-embodiment portability.}
UniStore advertises a single skill running across the G1 and H1 humanoids and the B2/Go2
quadrupeds; bodies with different morphologies, action spaces, and dynamics. This is the
cross-embodiment transfer problem (\S\ref{sec:transfer}) at deployment scale: without a
shared action interface \citep{openx} or explicit embodiment adaptation, a package tuned
for one platform has no guarantee of correctness on another. Formal portability, what must
hold for a skill to be certified on a new body, remains open.

\paragraph{Provenance and trust.}
A marketplace invites third-party, crowdsourced uploads, including raw motion-capture data
and code from unknown authors. Robot skills act on the physical world, so provenance
(who authored a skill, from what data, validated how) becomes a safety property, not merely
a metadata nicety. There is no accepted standard for signing, attesting, or auditing a
robot skill's origin.

\paragraph{Safety verification.}
Vendors promise that ``every skill is scanned, tested, and verified,'' but for a
code-as-policy skill this is undecidable in general and hard in practice: the skill's effect
depends on the scene it meets. Practical questions include which pre-conditions a skill must
declare, what sandboxed or simulated screening it must pass, and how run-time monitors
(cf.\ \emph{Code-as-Monitor} \citep{codeasmonitor}) can bound behaviour on unseen inputs.

\paragraph{Skill composition.}
Downloading two skills should ideally yield a third: ``make coffee'' plus ``clear the
table'' composed into a morning routine. Composition requires shared interfaces and a
calculus over pre-/post-conditions; today's motion packages are opaque and do not compose,
and code skills compose only within a single authoring system.

\paragraph{Portability standards and skill ontologies.}
Reuse at scale needs a shared vocabulary: declared pre-conditions, expected effects, and
embodiment-capability profiles, with relations such as \emph{requires}, \emph{provides}, and
\emph{substitutable-by}. Prior work on skill ontologies and hardware-level reusability
points the way, but no cross-vendor standard exists, which is what a genuine ecosystem would
require.

\paragraph{Local versus shared libraries.}
Finally, the field must reconcile two notions of ``skill library'' that this survey has kept
distinct: the \emph{self-built} library a robot grows from its own experience
(\S\ref{sec:code}) and the \emph{shared} library a robot downloads from a marketplace
(UniStore). The interesting regime is the hybrid, an agent that both contributes to and
draws from a commons, and its incentives, quality control, and feedback dynamics are
entirely unstudied.

\section{Limitations of this Survey}
\label{sec:limitations}

We state the boundaries of the survey explicitly. First, our organising axis, \emph{degree
of self-improvement}, is deliberately code-centric; it foregrounds systems that emit and
revise programs and treats weight-based policies (\S\ref{sec:vla}) and reinforcement-learning
skill discovery (\S\ref{sec:skill}) as context rather than as the object of the taxonomy.
A survey centred on representation learning or on real-world data collection would draw the
map differently. Second, the frontier we emphasise (\S3.1.5) is populated by very recent,
in some cases concurrent, systems; benchmark numbers across them are not directly
comparable, and we therefore report capabilities qualitatively rather than ranking methods.
Third, the field moves quickly: several 2025--2026 systems we cite were released while this
survey was being written, and the coverage should be read as a snapshot. Finally, the
``skill economy'' framing (\S\ref{sec:open}) is grounded in an emerging commercial trend
(one vendor's marketplace at the time of writing); we treat it as motivation for open
problems, not as a mature body of results.

\section{Future Directions}
\label{sec:future}

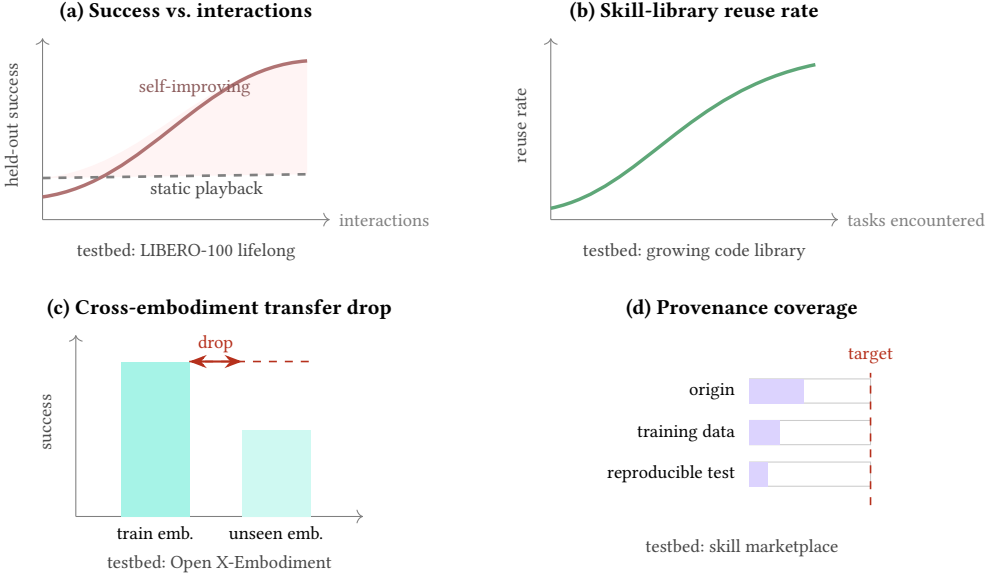
\begin{figure}[t]
\centering
\begin{tabular}{@{}c@{\hspace{9mm}}c@{}}
\begin{tikzpicture}[baseline]
  \draw[->,gray] (0,0)--(3.8,0) node[right,font=\scriptsize]{interactions};
  \draw[->,gray] (0,0)--(0,2.4);
  \node[rotate=90,font=\scriptsize,anchor=south,text=gray!60!black] at (-0.2,1.25){held-out success};
  \fill[secE!12] (0,0.55) .. controls (1.4,0.7) and (2.1,2.0) .. (3.5,2.1) -- (3.5,0.6) -- cycle;
  \draw[secE!70!black,line width=1.3pt] (0,0.3) .. controls (1.4,0.5) and (2.1,2.0) .. (3.5,2.1);
  \draw[gray,line width=1pt,dashed] (0,0.55)--(3.5,0.6);
  \node[secE!55!black,font=\scriptsize,anchor=west] at (1.15,1.75){self-improving};
  \node[gray!55!black,font=\scriptsize,anchor=west] at (1.3,0.4){static playback};
  \node[font=\footnotesize\bfseries] at (1.9,2.75){(a) Success vs.\ interactions};
  \node[font=\scriptsize,text=gray!55!black] at (1.9,-0.42){testbed: LIBERO-100 lifelong};
\end{tikzpicture}
&
\begin{tikzpicture}[baseline]
  \draw[->,gray] (0,0)--(3.8,0) node[right,font=\scriptsize]{tasks encountered};
  \draw[->,gray] (0,0)--(0,2.4);
  \node[rotate=90,font=\scriptsize,anchor=south,text=gray!60!black] at (-0.2,1.25){reuse rate};
  \draw[secC!70!black,line width=1.3pt] (0,0.15) .. controls (1.2,0.4) and (1.8,1.7) .. (3.5,2.05);
  \node[font=\footnotesize\bfseries] at (1.9,2.75){(b) Skill-library reuse rate};
  \node[font=\scriptsize,text=gray!55!black] at (1.9,-0.42){testbed: growing code library};
\end{tikzpicture}
\\[8mm]
\begin{tikzpicture}[baseline]
  \draw[->,gray] (0,0)--(3.8,0); \draw[->,gray] (0,0)--(0,2.4);
  \node[rotate=90,font=\scriptsize,anchor=south,text=gray!60!black] at (-0.2,1.25){success};
  \fill[cat-vla!55] (0.6,0) rectangle (1.5,2.05);
  \fill[cat-vla!30] (2.2,0) rectangle (3.1,1.15);
  \draw[BrickRed,{Stealth}-{Stealth},line width=0.8pt] (1.5,2.05)--(2.2,2.05);
  \node[BrickRed,font=\scriptsize,above] at (1.85,2.05){drop};
  \draw[BrickRed,dashed,line width=0.6pt] (1.5,2.05)--(3.1,2.05);
  \node[font=\scriptsize] at (1.05,-0.22){train emb.};
  \node[font=\scriptsize] at (2.65,-0.22){unseen emb.};
  \node[font=\footnotesize\bfseries] at (1.9,2.75){(c) Cross-embodiment transfer drop};
  \node[font=\scriptsize,text=gray!55!black] at (1.9,-0.62){testbed: Open X-Embodiment};
\end{tikzpicture}
&
\begin{tikzpicture}[baseline]
  \foreach [count=\i] \l/\f in {origin/0.45, {training data}/0.25, {reproducible test}/0.15}{
    \draw[gray!40] (2.0,{2.05-\i*0.55}) rectangle (3.6,{2.05-\i*0.55+0.32});
    \fill[cat-skill!60] (2.0,{2.05-\i*0.55}) rectangle ({2.0+1.6*\f},{2.05-\i*0.55+0.32});
    \node[left=2pt,font=\scriptsize] at (2.0,{2.05-\i*0.55+0.16}) {\l};
  }
  \draw[BrickRed,dashed,line width=0.6pt] (3.6,0.15)--(3.6,1.9) node[above,font=\scriptsize,text=BrickRed]{target};
  \node[font=\footnotesize\bfseries] at (1.9,2.75){(d) Provenance coverage};
  \node[font=\scriptsize,text=gray!55!black] at (1.9,-0.42){testbed: skill marketplace};
\end{tikzpicture}
\end{tabular}
\caption{\textbf{An actionable evaluation protocol for the future agenda} (\S\ref{sec:future};
Table~\ref{tab:future_metrics}). The panels are \emph{schematic}: they show the axes and the
\emph{target} shape of each measurement, not measured results. (a)~held-out success should rise with
autonomous interaction for a self-improving system while a static playback skill stays flat, on
LIBERO-style lifelong streams~\citep{libero}; (b)~a growing skill library should raise its reuse
rate; (c)~the drop in success from a training embodiment to an unseen one quantifies transfer, on
Open X-Embodiment-style splits~\citep{openx}; (d)~each distributed skill should carry a verifiable
origin, training-data statement, and reproducible test, a coverage today far below target for
marketplace skills~\citep{unistore2026}.}
\Description{Four schematic panels illustrating target measurement shapes, not real data. (a) A
rising, saturating curve labelled self-improving versus a flat dashed line labelled static playback,
on axes of held-out success against interactions, with the gap shaded; testbed LIBERO-100 lifelong.
(b) A rising curve of skill-library reuse rate against tasks encountered. (c) Two bars, a tall
train-embodiment bar and a shorter unseen-embodiment bar, with a red double arrow marking the drop;
testbed Open X-Embodiment. (d) Three partially filled bars for origin, training data, and
reproducible test, all short of a dashed target line; testbed skill marketplace.}
\label{fig:future_protocol}
\end{figure}

To make the agenda actionable rather than aspirational, we anchor it on four measurable
quantities, defined with concrete testbeds in Table~\ref{tab:future_metrics}: the
\emph{success-vs-interactions curve} (does competence rise with autonomous experience?), the
\emph{skill-library reuse rate} (does stored competence compound?), the \emph{cross-embodiment
transfer drop} (how much is lost when a skill changes bodies?), and the \emph{provenance check}
(can a distributed skill's origin and test record be verified?). The first two instantiate
naturally on LIBERO-style lifelong task streams~\citep{libero}; the third on held-out-embodiment
splits of Open X-Embodiment~\citep{openx}; the fourth on marketplace
listings~\citep{unistore2026}. Figure~\ref{fig:future_protocol} sketches the target shape each of
these measurements should take.

\begin{table*}[t]
\centering
\footnotesize
\setlength{\tabcolsep}{4pt}
\renewcommand{\arraystretch}{1.35}
\begin{tabular}{@{}p{0.17\textwidth} p{0.30\textwidth} p{0.27\textwidth} p{0.17\textwidth}@{}}
\toprule
\textbf{Metric} & \textbf{Definition} & \textbf{Instantiation (testbed)} & \textbf{Family stressed} \\
\midrule
Success-vs-interactions curve &
held-out success rate as a function of accumulated autonomous interactions; report the curve
and its area, not a single endpoint &
lifelong task streams in the style of LIBERO-100~\citep{libero}, fixed interaction budget &
\S\ref{sec:code} full loop; \S\ref{sec:vla} fine-tuning \\
Skill-library reuse rate &
fraction of new-task solutions that invoke at least one previously stored skill, and mean
invocations per stored skill &
growing code libraries~\citep{voyager} evaluated across LIBERO-style task suites~\citep{libero} &
\S\ref{sec:skill} libraries; \S\ref{sec:code} memory rung \\
Cross-embodiment transfer drop &
difference in success between the training embodiment and an unseen one, at matched task and
budget &
held-out-embodiment splits of Open X-Embodiment~\citep{openx} (22 embodiments) &
\S\ref{sec:vla}; \S\ref{sec:transfer} \\
Provenance check &
fraction of distributed skills carrying a verifiable origin, training-data statement, and
reproducible test record &
audits of marketplace listings~\citep{unistore2026} against a declared provenance schema &
market skills (\S\ref{sec:open}) \\
\bottomrule
\end{tabular}
\caption{\textbf{An actionable protocol for the future agenda}: four measurable quantities, their
definitions, and concrete testbeds. The first two instantiate LIBERO-style lifelong
learning~\citep{libero}; the third instantiates Open X-Embodiment-style multi-robot
transfer~\citep{openx}; the fourth targets the skill marketplaces of \S\ref{sec:open}.}
\label{tab:future_metrics}
\end{table*}

Three directions follow directly from the map, each now with its success measure.
\textbf{(1) Standardised evaluation of self-improvement.} The systems on the top rung
(\S3.1.5) each report improvement on their own splits; the field lacks a shared protocol that
plots the success-vs-interactions curve and the skill-library reuse rate under a common interaction
budget. A benchmark that publishes both curves per system, in the way LIBERO standardised
lifelong manipulation~\citep{libero}, would make ``self-improvement'' a reported number rather
than a claim. \textbf{(2) Adaptation as a first-class marketplace primitive.} The open problems
of \S\ref{sec:open} suggest a research programme in which a downloaded skill is not a frozen
artefact but a starting point that the \S\ref{sec:code} loop specialises to the local robot and
environment; its success measures are the transfer drop before versus after on-device
adaptation, and the provenance check on what was downloaded. \textbf{(3) Bridging the RL and
code flavours of skills.} \S\ref{sec:skill} shows that skills are learned either as latent RL
policies or as retrievable code; a unifying interface, code that wraps, calls, and refines
learned policies, and learned policies distilled back into inspectable code, would let the
self-improvement machinery of \S\ref{sec:code} operate over both, and its progress is legible in
the reuse rate crossing family boundaries. We see the combination of a standard self-improvement
benchmark with an adaptation-centric marketplace as the most likely path from today's static
skill stores to genuinely capable, compounding robots.

\section{Conclusion}
\label{sec:conclusion}

Robot learning is organising itself around a single question: should competence be shipped
as \emph{weights} or as \emph{skills}? This survey took that question as its axis. We mapped
the field into six branches (\S\ref{sec:taxonomy}), positioned end-to-end vision-language-
action models as the weights-based foil (\S\ref{sec:vla}), and gave the code-as-policy camp
its own deep-dive (\S\ref{sec:code}) organised by \emph{degree of self-improvement}: from
one-shot program synthesis, through within-task repair, persistent skill memory, and
population search, to the open-ended loop that combines all three. That top cell, feedback
$+$ memory $+$ search, is the least populated and least surveyed region of the field, and
it is where systems such as ASPIRE, ENPIRE, and RoboClaw now sit. We complemented this with a map of
how skill \emph{repertoires} are built (\S\ref{sec:skill}), showing that skills come in an
RL flavour and a code flavour of which only the latter self-improves without gradients, and
we connected the academic taxonomy to the nascent skill economy (\S\ref{sec:open}), where
commercial marketplaces already distribute skills but ship only static playback. The gap
between that static distribution and genuine on-device adaptation is, we argue, the defining
research opportunity for physical AI, and the self-improvement ladder of \S\ref{sec:code} offers a structured path toward closing it.

\bibliographystyle{ACM-Reference-Format}
\bibliography{refs}

\clearpage
\appendix
\section{The Broader Robot-Learning Landscape}
\label{sec:landscape}

The taxonomy of Figure~\ref{fig:taxonomy_main} foregrounds 77 systems chosen, by the placement
criteria of \S\ref{sec:method}, to populate a single analytical axis, weights versus skills, rather
than to enumerate the field. To situate those systems
within the wider research landscape, and to make the survey's coverage inspectable at a glance,
Table~\ref{tab:landscape} catalogues a further set of representative works: end-to-end and generalist
policies, imitation and diffusion policies, language-model planning and task-and-motion planning,
reward and data synthesis, unsupervised skill discovery and hierarchical reinforcement learning, world
models and model-based control, representation learning and offline reinforcement learning,
manipulation, dexterity, and locomotion, and the datasets, benchmarks, and simulators that support
them. Together with the 77 systems detailed in the main text
(Tables~\ref{tab:compare_main}--\ref{tab:compare_34}), the works surveyed number more than three
hundred. The list is representative rather than exhaustive; each entry is a real, individually
citeable paper, grouped by area and ordered by year. This set is characterised in the main text
(\S\ref{sec:corpus_gallery}): Figure~\ref{fig:corpus_overview} summarises the corpus at a glance,
Figure~\ref{fig:corpus_dist} profiles it by publication year, evaluation domain, learning signal,
and embodiment, and the visual galleries (Figures~\ref{fig:robots_canvas} and~\ref{fig:plots_canvas})
show the surveyed systems; Table~\ref{tab:canvas_provenance} lists, for each gallery panel, the
source paper and the figure number within it, so that no image is used without a traceable,
clickable origin.

\begin{table*}[t]\centering\small
\caption{\textbf{Per-panel provenance for Figures~\ref{fig:robots_canvas},~\ref{fig:plots_canvas}, and~\ref{fig:corpus_collage}.} Every panel is cropped from the listed figure of the cited paper; the same links are embedded, clickable, in the figures themselves. $\dagger$ marks a panel whose source figure is a redrawn/replicated variant or appears identically more than once in the paper (matched by content). The corpus-at-a-glance collage (Figure~\ref{fig:corpus_collage}) reuses the robot panels and is not relisted.}
\label{tab:canvas_provenance}
\begin{tabular}{@{}llll@{}}
\toprule
System & Paper & Source figure & Source file \\
\midrule
\multicolumn{4}{@{}l}{\textit{Figure~\ref{fig:robots_canvas}: robots (27 panels)}} \\
RT-1 & \href{https://arxiv.org/abs/2212.06817}{arXiv:2212.06817} & \href{https://arxiv.org/abs/2212.06817}{Fig. 1} & rt1\_teaser\_tasks.png \\
RT-2 & \href{https://arxiv.org/abs/2307.15818}{arXiv:2307.15818} & \href{https://arxiv.org/abs/2307.15818}{Fig. 2} & RT2-capabilities-dm.png \\
OpenVLA & \href{https://arxiv.org/abs/2406.09246}{arXiv:2406.09246} & \href{https://arxiv.org/abs/2406.09246}{Fig. 11} & droid\_wipe\_task.jpeg \\
Octo & \href{https://arxiv.org/abs/2405.12213}{arXiv:2405.12213} & \href{https://arxiv.org/abs/2405.12213}{Fig. 4}\,$\dagger$ & pdf\_im-164.png \\
$\pi_0$ & \href{https://arxiv.org/abs/2410.24164}{arXiv:2410.24164} & \href{https://arxiv.org/abs/2410.24164}{Fig. 2} & fig2\_final.jpeg \\
CogACT & \href{https://arxiv.org/abs/2411.19650}{arXiv:2411.19650} & \href{https://arxiv.org/abs/2411.19650}{Fig. II (suppl.)} & franka\_robot\_with\_label.png \\
ProgPrompt & \href{https://arxiv.org/abs/2209.11302}{arXiv:2209.11302} & \href{https://arxiv.org/abs/2209.11302}{Fig. 1} & pdf\_im-000.png \\
VoxPoser & \href{https://arxiv.org/abs/2307.05973}{arXiv:2307.05973} & \href{https://arxiv.org/abs/2307.05973}{Fig. 1} & pdf\_im-005.png \\
Code-as-Monitor & \href{https://arxiv.org/abs/2412.04455}{arXiv:2412.04455} & \href{https://arxiv.org/abs/2412.04455}{Fig. 1} & pdf\_im-002.png \\
REFLECT & \href{https://arxiv.org/abs/2306.15724}{arXiv:2306.15724} & \href{https://arxiv.org/abs/2306.15724}{Fig. 1} & pdf\_im-029.png \\
LOTUS & \href{https://arxiv.org/abs/2311.02058}{arXiv:2311.02058} & \href{https://arxiv.org/abs/2311.02058}{Fig. 1} & pdf\_im-014.png \\
CaP-X & \href{https://arxiv.org/abs/2603.22435}{arXiv:2603.22435} & \href{https://arxiv.org/abs/2603.22435}{Fig. 21} & seq4.png \\
UniSkill & \href{https://arxiv.org/abs/2603.02623}{arXiv:2603.02623} & \href{https://arxiv.org/abs/2603.02623}{Fig. 4} & pdf\_im-121.png \\
Gemini Robotics & \href{https://arxiv.org/abs/2503.20020}{arXiv:2503.20020} & \href{https://arxiv.org/abs/2503.20020}{Fig. 10} & mv1.jpeg \\
TidyBot & \href{https://arxiv.org/abs/2305.05658}{arXiv:2305.05658} & \href{https://arxiv.org/abs/2305.05658}{Fig. 4} & IMG\_5028.jpg \\
DrEureka & \href{https://arxiv.org/abs/2406.01967}{arXiv:2406.01967} & \href{https://arxiv.org/abs/2406.01967}{Fig. 4} & quadruped\_terrains.png \\
DrEureka & \href{https://arxiv.org/abs/2406.01967}{arXiv:2406.01967} & \href{https://arxiv.org/abs/2406.01967}{Fig. 2} & robots.png \\
ENPIRE & \href{https://arxiv.org/abs/2606.19980}{arXiv:2606.19980} & \href{https://arxiv.org/abs/2606.19980}{Fig. 11} & four\_camera\_setup.jpg \\
CrossFormer & \href{https://arxiv.org/abs/2408.11812}{arXiv:2408.11812} & \href{https://arxiv.org/abs/2408.11812}{Fig. 4} & pdf\_im-099.png \\
CrossFormer & \href{https://arxiv.org/abs/2408.11812}{arXiv:2408.11812} & \href{https://arxiv.org/abs/2408.11812}{Fig. 4} & pdf\_im-098.png \\
Meta-World & \href{https://arxiv.org/abs/1910.10897}{arXiv:1910.10897} & \href{https://arxiv.org/abs/1910.10897}{Fig. 1} & pdf\_im-000.png \\
robosuite & \href{https://arxiv.org/abs/2009.12293}{arXiv:2009.12293} & \href{https://arxiv.org/abs/2009.12293}{env fig. (unnum., p.14)}\,$\dagger$ & env\_door\_v15.png \\
Instruct2Act & \href{https://arxiv.org/abs/2305.11176}{arXiv:2305.11176} & \href{https://arxiv.org/abs/2305.11176}{Fig. 1} & pdf\_im-003.png \\
Prompt2Walk & \href{https://arxiv.org/abs/2309.09969}{arXiv:2309.09969} & \href{https://arxiv.org/abs/2309.09969}{Fig. 1} & pdf\_im-062.png \\
DoReMi & \href{https://arxiv.org/abs/2307.00329v3}{arXiv:2307.00329v3} & \href{https://arxiv.org/abs/2307.00329v3}{Fig. 11b} & ft1.png \\
DIAYN & \href{https://arxiv.org/abs/1802.06070}{arXiv:1802.06070} & \href{https://arxiv.org/abs/1802.06070}{Fig. 3} & skills.jpg \\
Mirage & \href{https://arxiv.org/abs/2402.19249}{arXiv:2402.19249} & \href{https://arxiv.org/abs/2402.19249}{Fig. 2} & Sim\_Tasks\_Robots\_Fig.png \\
\midrule
\multicolumn{4}{@{}l}{\textit{Figure~\ref{fig:plots_canvas}: results (20 panels)}} \\
RT-2 & \href{https://arxiv.org/abs/2307.15818}{arXiv:2307.15818} & \href{https://arxiv.org/abs/2307.15818}{Fig. 6a} & rt2\_emergent\_dm.png \\
Code-as-Policies & \href{https://arxiv.org/abs/2209.07753}{arXiv:2209.07753} & \href{https://arxiv.org/abs/2209.07753}{Fig. 4} & generalization\_types\_flat.png \\
Demo2Code & \href{https://arxiv.org/abs/2305.16744}{arXiv:2305.16744} & \href{https://arxiv.org/abs/2305.16744}{Fig. 13} & code\_ablation\_v4.png \\
Eureka & \href{https://arxiv.org/abs/2310.12931}{arXiv:2310.12931} & \href{https://arxiv.org/abs/2310.12931}{Fig. 13} & dexterity\_bar\_chart.png \\
RoboScript & \href{https://arxiv.org/abs/2402.14623}{arXiv:2402.14623} & \href{https://arxiv.org/abs/2402.14623}{Fig. 8} & ablation\_drawer\_place.png \\
$\pi_0$ & \href{https://arxiv.org/abs/2410.24164}{arXiv:2410.24164} & \href{https://arxiv.org/abs/2410.24164}{Fig. 13} & complex\_finetune.png \\
Text2Motion & \href{https://arxiv.org/abs/2303.12153}{arXiv:2303.12153} & \href{https://arxiv.org/abs/2303.12153}{Fig. 5} & Figure\_5.jpg \\
DrEureka & \href{https://arxiv.org/abs/2406.01967}{arXiv:2406.01967} & \href{https://arxiv.org/abs/2406.01967}{Fig. 8} & quadruped\_training\_curves.png \\
Eureka & \href{https://arxiv.org/abs/2310.12931}{arXiv:2310.12931} & \href{https://arxiv.org/abs/2310.12931}{Fig. 10} & bidex\_training\_curves.png \\
ManiSkill2 & \href{https://arxiv.org/abs/2302.04659}{arXiv:2302.04659} & \href{https://arxiv.org/abs/2302.04659}{Fig. 3} & CNNE.png \\
$\pi_{0.5}$ & \href{https://arxiv.org/abs/2504.16054}{arXiv:2504.16054} & \href{https://arxiv.org/abs/2504.16054}{Fig. 9} & env\_scaling\_results0.png \\
Voyager & \href{https://arxiv.org/abs/2305.16291}{arXiv:2305.16291} & \href{https://arxiv.org/abs/2305.16291}{Fig. A.4} & model\_variations.png \\
DIAYN & \href{https://arxiv.org/abs/1802.06070}{arXiv:1802.06070} & \href{https://arxiv.org/abs/1802.06070}{Fig. 4} & cheetah\_entropy.png \\
DADS & \href{https://arxiv.org/abs/1907.01657}{arXiv:1907.01657} & \href{https://arxiv.org/abs/1907.01657}{Fig. 8} & hierarchical\_control\_usl\_fix4.png \\
$\pi_0$ & \href{https://arxiv.org/abs/2410.24164}{arXiv:2410.24164} & \href{https://arxiv.org/abs/2410.24164}{Fig. 4} & combined-robot-allocation-chart.png \\
SkiMo & \href{https://arxiv.org/abs/2207.07560}{arXiv:2207.07560} & \href{https://arxiv.org/abs/2207.07560}{Fig. 9} & ours\_coverage.png \\
OPAL & \href{https://arxiv.org/abs/2010.13611}{arXiv:2010.13611} & \href{https://arxiv.org/abs/2010.13611}{Fig. 1} & antmaze\_medium.png \\
LRLL & \href{https://arxiv.org/abs/2406.18746}{arXiv:2406.18746} & \href{https://arxiv.org/abs/2406.18746}{Fig. 3}\,$\dagger$ & tsne231.drawio.png \\
LSD & \href{https://arxiv.org/abs/2202.00914}{arXiv:2202.00914} & \href{https://arxiv.org/abs/2202.00914}{Fig. 1b}\,$\dagger$ & pdf\_im-001.png \\
OPAL & \href{https://arxiv.org/abs/2010.13611}{arXiv:2010.13611} & \href{https://arxiv.org/abs/2010.13611}{Fig. 3} & cql\_large\_heat.png \\
\bottomrule
\end{tabular}
\end{table*}

{\scriptsize
\setlength{\tabcolsep}{4pt}
\renewcommand{\arraystretch}{1.1}
\begin{longtable}{@{}p{0.27\textwidth} c l l c l c@{}}
\caption{\textbf{The broader robot-learning landscape: 225 representative works, compared on six axes.} Beyond the 77 systems placed in the taxonomy (Figure~\ref{fig:taxonomy_main}, Tables~\ref{tab:compare_main}--\ref{tab:compare_34}), this table catalogues the wider field, grouped into eleven areas and sub-grouped by family. Columns: \textbf{Year}, \textbf{Ships} (what the method contributes), \textbf{Learn} (primary learning signal: IL, RL, offline-RL, LLM, self-sup., model-based), \textbf{Eval} (\dreal/\dsim/\dgame/offline), \textbf{Embod.} (embodiment), and run-time \textbf{S.I.} (self-improvement: \yy\ yes, \pp\ partial, \nn\ no). A \dm\ marks not-applicable or unverified. Every entry is a real, individually citeable paper.}\label{tab:landscape}\\
\toprule
\textbf{System / work} & \textbf{Year} & \textbf{Ships} & \textbf{Learn} & \textbf{Eval} & \textbf{Embod.} & \textbf{S.I.}\\
\midrule
\endfirsthead
\multicolumn{7}{@{}l}{\scriptsize\emph{Table~\ref{tab:landscape} (continued)}}\\
\toprule
\textbf{System / work} & \textbf{Year} & \textbf{Ships} & \textbf{Learn} & \textbf{Eval} & \textbf{Embod.} & \textbf{S.I.}\\
\midrule
\endhead
\bottomrule
\endlastfoot
\multicolumn{7}{@{}l}{\rule{0pt}{2.6ex}\textbf{\textsf{End-to-end \& generalist policies}}~\textcolor{gray}{(28)}}\\[1pt]
\multicolumn{7}{@{}l}{\hspace{0.6em}\emph{Vision-language-action}}\\
\hspace{0.6em}\textbf{3D-VLA}~\cite{vla3d} & 2024 & \textsf{world-model} & \textsf{LLM} & \dsim & \textsf{arm} & \nn\\
\hspace{0.6em}\textbf{ECoT}~\cite{ecot} & 2024 & \textsf{policy} & \textsf{IL} & \dreal & \textsf{arm} & \nn\\
\hspace{0.6em}\textbf{LLaRA}~\cite{llara} & 2024 & \textsf{policy} & \textsf{LLM} & \dsim & \textsf{arm} & \nn\\
\hspace{0.6em}\textbf{ManipLLM}~\cite{manipllm} & 2024 & \textsf{policy} & \textsf{LLM} & \dsim & \textsf{arm} & \nn\\
\hspace{0.6em}\textbf{RoboMamba}~\cite{robomamba} & 2024 & \textsf{policy} & \textsf{LLM} & \dsim & \textsf{arm} & \nn\\
\hspace{0.6em}\textbf{RoboPoint}~\cite{robopoint} & 2024 & \textsf{plan} & \textsf{LLM} & \dreal & \textsf{multi} & \nn\\
\hspace{0.6em}\textbf{TinyVLA}~\cite{tinyvla} & 2024 & \textsf{policy} & \textsf{IL} & \dreal & \textsf{arm} & \nn\\
\hspace{0.6em}\textbf{TraceVLA}~\cite{tracevla} & 2024 & \textsf{policy} & \textsf{IL} & \dsim & \textsf{arm} & \nn\\
\hspace{0.6em}\textbf{DexVLA}~\cite{dexvla} & 2025 & \textsf{policy} & \textsf{IL} & \dreal & \textsf{multi} & \nn\\
\hspace{0.6em}\textbf{FAST}~\cite{fast} & 2025 & \textsf{repr} & \textsf{IL} & \dreal & \textsf{arm} & \nn\\
\hspace{0.6em}\textbf{OpenVLA-OFT}~\cite{openvlaoft} & 2025 & \textsf{policy} & \textsf{IL} & \dsim & \textsf{arm} & \nn\\
\multicolumn{7}{@{}l}{\hspace{0.6em}\emph{Generalist / multi-task}}\\
\hspace{0.6em}\textbf{QT-Opt}~\cite{qtopt} & 2018 & \textsf{policy} & \textsf{RL} & \dreal & \textsf{arm} & \pp\\
\hspace{0.6em}\textbf{MT-Opt}~\cite{mtopt} & 2021 & \textsf{policy} & \textsf{RL} & \dreal & \textsf{arm} & \pp\\
\hspace{0.6em}\textbf{Gato}~\cite{gato} & 2022 & \textsf{policy} & \textsf{IL} & \dsim & \textsf{multi} & \nn\\
\hspace{0.6em}\textbf{PaLM-E}~\cite{palme} & 2023 & \textsf{plan} & \textsf{LLM} & \dreal & \textsf{multi} & \nn\\
\hspace{0.6em}\textbf{Q-Transformer}~\cite{qtransformer} & 2023 & \textsf{policy} & \textsf{offline-RL} & \dreal & \textsf{arm} & \nn\\
\hspace{0.6em}\textbf{UniPi}~\cite{unipi} & 2023 & \textsf{policy} & \textsf{model-based} & \dsim & \textsf{arm} & \nn\\
\hspace{0.6em}\textbf{VIMA}~\cite{vima} & 2023 & \textsf{policy} & \textsf{IL} & \dsim & \textsf{arm} & \nn\\
\hspace{0.6em}\textbf{GR-2}~\cite{gr2} & 2024 & \textsf{policy} & \textsf{IL} & \dreal & \textsf{arm} & \nn\\
\hspace{0.6em}\textbf{LEO}~\cite{leo} & 2024 & \textsf{policy} & \textsf{LLM} & \dsim & \textsf{multi} & \nn\\
\hspace{0.6em}\textbf{Magma}~\cite{magma} & 2025 & \textsf{weights} & \textsf{LLM} & \dsim & \textsf{arm} & \nn\\
\multicolumn{7}{@{}l}{\hspace{0.6em}\emph{Video / action pretraining}}\\
\hspace{0.6em}\textbf{ATM}~\cite{atm} & 2024 & \textsf{policy} & \textsf{self-sup} & \dsim & \textsf{arm} & \nn\\
\hspace{0.6em}\textbf{GR-1}~\cite{gr1} & 2024 & \textsf{policy} & \textsf{IL} & \dsim & \textsf{arm} & \nn\\
\hspace{0.6em}\textbf{HPT}~\cite{hpt} & 2024 & \textsf{repr} & \textsf{IL} & \dsim & \textsf{multi} & \nn\\
\hspace{0.6em}\textbf{VPP}~\cite{vpp} & 2024 & \textsf{policy} & \textsf{IL} & \dsim & \textsf{arm} & \nn\\
\hspace{0.6em}\textbf{Cosmos}~\cite{cosmos} & 2025 & \textsf{world-model} & \textsf{self-sup} & \dm & \dm & \nn\\
\hspace{0.6em}\textbf{LAPA}~\cite{lapa} & 2025 & \textsf{policy} & \textsf{self-sup} & \dsim & \textsf{arm} & \nn\\
\hspace{0.6em}\textbf{Seer}~\cite{seer} & 2025 & \textsf{policy} & \textsf{IL} & \dsim & \textsf{arm} & \nn\\
\addlinespace[2pt]
\multicolumn{7}{@{}l}{\rule{0pt}{2.6ex}\textbf{\textsf{Imitation \& diffusion policies}}~\textcolor{gray}{(21)}}\\[1pt]
\multicolumn{7}{@{}l}{\hspace{0.6em}\emph{Imitation learning}}\\
\hspace{0.6em}\textbf{BC-Z}~\cite{bcz} & 2021 & \textsf{policy} & \textsf{IL} & \dreal & \textsf{arm} & \nn\\
\hspace{0.6em}\textbf{Implicit BC}~\cite{ibc} & 2021 & \textsf{policy} & \textsf{IL} & \dsim & \textsf{arm} & \nn\\
\hspace{0.6em}\textbf{robomimic}~\cite{robomimic} & 2021 & \textsf{bench} & \textsf{IL} & \dsim & \textsf{arm} & \nn\\
\hspace{0.6em}\textbf{PerAct}~\cite{peract} & 2022 & \textsf{policy} & \textsf{IL} & \dsim & \textsf{arm} & \nn\\
\hspace{0.6em}\textbf{ACT/ALOHA}~\cite{act} & 2023 & \textsf{policy} & \textsf{IL} & \dreal & \textsf{arm} & \nn\\
\hspace{0.6em}\textbf{MimicGen}~\cite{mimicgen} & 2023 & \textsf{data} & \textsf{IL} & \dsim & \textsf{arm} & \nn\\
\hspace{0.6em}\textbf{RVT}~\cite{rvt} & 2023 & \textsf{policy} & \textsf{IL} & \dsim & \textsf{arm} & \nn\\
\hspace{0.6em}\textbf{Mobile ALOHA}~\cite{mobilealoha} & 2024 & \textsf{policy} & \textsf{IL} & \dreal & \textsf{mobile} & \nn\\
\hspace{0.6em}\textbf{RoboAgent}~\cite{roboagent} & 2024 & \textsf{policy} & \textsf{IL} & \dreal & \textsf{arm} & \nn\\
\hspace{0.6em}\textbf{RVT-2}~\cite{rvt2} & 2024 & \textsf{policy} & \textsf{IL} & \dsim & \textsf{arm} & \nn\\
\hspace{0.6em}\textbf{UMI}~\cite{umi} & 2024 & \textsf{data} & \textsf{IL} & \dreal & \textsf{arm} & \nn\\
\multicolumn{7}{@{}l}{\hspace{0.6em}\emph{Diffusion policies}}\\
\hspace{0.6em}\textbf{Diffuser}~\cite{diffuser} & 2022 & \textsf{plan} & \textsf{model-based} & \textcolor{gray!70}{\textsf{offline}} & \dm & \nn\\
\hspace{0.6em}\textbf{BESO}~\cite{beso} & 2023 & \textsf{policy} & \textsf{IL} & \dsim & \textsf{arm} & \nn\\
\hspace{0.6em}\textbf{Decision Diffuser}~\cite{decisiondiffuser} & 2023 & \textsf{policy} & \textsf{offline-RL} & \textcolor{gray!70}{\textsf{offline}} & \dm & \nn\\
\hspace{0.6em}\textbf{Diffusion Policy}~\cite{diffusionpolicy} & 2023 & \textsf{policy} & \textsf{IL} & \dsim & \textsf{arm} & \nn\\
\hspace{0.6em}\textbf{3D Diffuser Actor}~\cite{diffuseractor3d} & 2024 & \textsf{policy} & \textsf{IL} & \dsim & \textsf{arm} & \nn\\
\hspace{0.6em}\textbf{DP3}~\cite{dp3} & 2024 & \textsf{policy} & \textsf{IL} & \dsim & \textsf{arm} & \nn\\
\hspace{0.6em}\textbf{Equivariant Diffusion Policy}~\cite{equidiff} & 2024 & \textsf{policy} & \textsf{IL} & \dsim & \textsf{arm} & \nn\\
\hspace{0.6em}\textbf{MDT}~\cite{mdt} & 2024 & \textsf{policy} & \textsf{IL} & \dsim & \textsf{arm} & \nn\\
\hspace{0.6em}\textbf{RDT-1B}~\cite{rdt1b} & 2024 & \textsf{weights} & \textsf{IL} & \dreal & \textsf{arm} & \nn\\
\hspace{0.6em}\textbf{SuSIE}~\cite{susie} & 2024 & \textsf{policy} & \textsf{IL} & \dsim & \textsf{arm} & \nn\\
\addlinespace[2pt]
\multicolumn{7}{@{}l}{\rule{0pt}{2.6ex}\textbf{\textsf{LLM/VLM planning \& task-and-motion}}~\textcolor{gray}{(30)}}\\[1pt]
\multicolumn{7}{@{}l}{\hspace{0.6em}\emph{LLM task planning}}\\
\hspace{0.6em}\textbf{COWP}~\cite{cowp} & 2022 & \textsf{plan} & \textsf{LLM} & \dsim & \textsf{mobile} & \pp\\
\hspace{0.6em}\textbf{ReAct}~\cite{react} & 2022 & \textsf{plan} & \textsf{LLM} & \dgame & \dm & \pp\\
\hspace{0.6em}\textbf{SayCan}~\cite{saycan} & 2022 & \textsf{plan} & \textsf{LLM} & \dreal & \textsf{mobile} & \nn\\
\hspace{0.6em}\textbf{Socratic Models}~\cite{socraticmodels} & 2022 & \textsf{plan} & \textsf{LLM} & \dreal & \textsf{arm} & \nn\\
\hspace{0.6em}\textbf{Zero-Shot Planners}~\cite{zeroshotplanners} & 2022 & \textsf{plan} & \textsf{LLM} & \dsim & \dm & \nn\\
\hspace{0.6em}\textbf{DEPS}~\cite{deps} & 2023 & \textsf{plan} & \textsf{LLM} & \dgame & \dm & \pp\\
\hspace{0.6em}\textbf{GLAM}~\cite{glam} & 2023 & \textsf{policy} & \textsf{RL} & \dgame & \dm & \nn\\
\hspace{0.6em}\textbf{Grounded Decoding}~\cite{grounddecoding} & 2023 & \textsf{plan} & \textsf{LLM} & \dreal & \textsf{mobile} & \nn\\
\hspace{0.6em}\textbf{ISR-LLM}~\cite{isrllm} & 2023 & \textsf{plan} & \textsf{LLM} & \dsim & \dm & \pp\\
\hspace{0.6em}\textbf{ITP}~\cite{interactivetaskplanning} & 2023 & \textsf{plan} & \textsf{LLM} & \dreal & \textsf{arm} & \pp\\
\hspace{0.6em}\textbf{KnowNo}~\cite{knowno} & 2023 & \textsf{plan} & \textsf{LLM} & \dreal & \textsf{arm} & \nn\\
\hspace{0.6em}\textbf{LLM-Planner}~\cite{llmplanner} & 2023 & \textsf{plan} & \textsf{LLM} & \dsim & \textsf{mobile} & \pp\\
\hspace{0.6em}\textbf{Reflexion}~\cite{reflexion} & 2023 & \textsf{plan} & \textsf{LLM} & \dgame & \dm & \yy\\
\hspace{0.6em}\textbf{RoCo}~\cite{roco} & 2023 & \textsf{plan} & \textsf{LLM} & \dsim & \textsf{multi} & \pp\\
\hspace{0.6em}\textbf{SayPlan}~\cite{sayplan} & 2023 & \textsf{plan} & \textsf{LLM} & \dreal & \textsf{mobile} & \pp\\
\hspace{0.6em}\textbf{Self-Refine}~\cite{selfrefine} & 2023 & \textsf{plan} & \textsf{LLM} & \dm & \dm & \yy\\
\hspace{0.6em}\textbf{SMART-LLM}~\cite{smartllm} & 2023 & \textsf{plan} & \textsf{LLM} & \dsim & \textsf{multi} & \nn\\
\hspace{0.6em}\textbf{Tree-Planner}~\cite{treeplanner} & 2023 & \textsf{plan} & \textsf{LLM} & \dsim & \dm & \pp\\
\hspace{0.6em}\textbf{ReAd}~\cite{readmarl} & 2024 & \textsf{plan} & \textsf{LLM} & \dsim & \textsf{multi} & \pp\\
\multicolumn{7}{@{}l}{\hspace{0.6em}\emph{LLM task-and-motion}}\\
\hspace{0.6em}\textbf{AutoTAMP}~\cite{autotamp} & 2023 & \textsf{plan} & \textsf{LLM} & \dsim & \dm & \pp\\
\hspace{0.6em}\textbf{Lang2LTL}~\cite{lang2ltl} & 2023 & \textsf{plan} & \textsf{LLM} & \dreal & \textsf{mobile} & \nn\\
\hspace{0.6em}\textbf{LLM+P}~\cite{llmp} & 2023 & \textsf{plan} & \textsf{LLM} & \dm & \dm & \nn\\
\hspace{0.6em}\textbf{LLM-GROP}~\cite{llmgrop} & 2023 & \textsf{plan} & \textsf{LLM} & \dsim & \textsf{mobile} & \nn\\
\hspace{0.6em}\textbf{NL2TL}~\cite{nl2tl} & 2023 & \textsf{plan} & \textsf{LLM} & \textcolor{gray!70}{\textsf{offline}} & \dm & \nn\\
\hspace{0.6em}\textbf{DELTA}~\cite{delta} & 2024 & \textsf{plan} & \textsf{LLM} & \dsim & \textsf{mobile} & \nn\\
\hspace{0.6em}\textbf{LLM3}~\cite{llm3} & 2024 & \textsf{plan} & \textsf{LLM} & \dsim & \textsf{arm} & \pp\\
\multicolumn{7}{@{}l}{\hspace{0.6em}\emph{Code / prompt as policy}}\\
\hspace{0.6em}\textbf{LM Traj Generators}~\cite{zeroshottraj} & 2023 & \textsf{plan} & \textsf{LLM} & \dreal & \textsf{arm} & \pp\\
\hspace{0.6em}\textbf{MOKA}~\cite{moka} & 2024 & \textsf{plan} & \textsf{LLM} & \dreal & \textsf{arm} & \nn\\
\hspace{0.6em}\textbf{PIVOT}~\cite{pivot} & 2024 & \textsf{plan} & \textsf{LLM} & \dreal & \textsf{mobile} & \pp\\
\hspace{0.6em}\textbf{ReKep}~\cite{rekep} & 2024 & \textsf{code} & \textsf{LLM} & \dreal & \textsf{arm} & \pp\\
\addlinespace[2pt]
\multicolumn{7}{@{}l}{\rule{0pt}{2.6ex}\textbf{\textsf{LLM reward synthesis \& data generation}}~\textcolor{gray}{(17)}}\\[1pt]
\multicolumn{7}{@{}l}{\hspace{0.6em}\emph{LLM reward synthesis}}\\
\hspace{0.6em}\textbf{ELLM}~\cite{ellm} & 2023 & \textsf{reward} & \textsf{RL} & \dgame & \dm & \nn\\
\hspace{0.6em}\textbf{LAMP}~\cite{lamp} & 2023 & \textsf{reward} & \textsf{RL} & \dsim & \textsf{arm} & \nn\\
\hspace{0.6em}\textbf{Motif}~\cite{motif} & 2023 & \textsf{reward} & \textsf{RL} & \dgame & \dm & \nn\\
\hspace{0.6em}\textbf{Read and Reap}~\cite{readreap} & 2023 & \textsf{reward} & \textsf{RL} & \dgame & \dm & \nn\\
\hspace{0.6em}\textbf{VLM-RM}~\cite{vlmrm} & 2023 & \textsf{reward} & \textsf{RL} & \dsim & \textsf{humanoid} & \nn\\
\hspace{0.6em}\textbf{Code as Reward}~\cite{codeasreward} & 2024 & \textsf{reward} & \textsf{RL} & \dsim & \dm & \nn\\
\hspace{0.6em}\textbf{RL-VLM-F}~\cite{rlvlmf} & 2024 & \textsf{reward} & \textsf{RL} & \dsim & \textsf{arm} & \nn\\
\multicolumn{7}{@{}l}{\hspace{0.6em}\emph{Generative data / augmentation}}\\
\hspace{0.6em}\textbf{CACTI}~\cite{cacti} & 2022 & \textsf{data} & \textsf{IL} & \dreal & \textsf{arm} & \nn\\
\hspace{0.6em}\textbf{DIAL}~\cite{dial} & 2022 & \textsf{data} & \textsf{IL} & \dreal & \textsf{mobile} & \nn\\
\hspace{0.6em}\textbf{GenAug}~\cite{genaug} & 2023 & \textsf{data} & \textsf{IL} & \dreal & \textsf{arm} & \nn\\
\hspace{0.6em}\textbf{GenSim}~\cite{gensim} & 2023 & \textsf{data} & \textsf{IL} & \dsim & \textsf{arm} & \nn\\
\hspace{0.6em}\textbf{ROSIE}~\cite{rosie} & 2023 & \textsf{data} & \textsf{IL} & \dreal & \textsf{mobile} & \nn\\
\hspace{0.6em}\textbf{Scaling Up Distilling Down}~\cite{scalingup} & 2023 & \textsf{policy} & \textsf{IL} & \dreal & \textsf{arm} & \nn\\
\hspace{0.6em}\textbf{BBSEA}~\cite{bbsea} & 2024 & \textsf{skill} & \textsf{IL} & \dsim & \textsf{arm} & \pp\\
\hspace{0.6em}\textbf{DexMimicGen}~\cite{dexmimicgen} & 2024 & \textsf{data} & \textsf{IL} & \dsim & \textsf{humanoid} & \nn\\
\hspace{0.6em}\textbf{GenSim2}~\cite{gensim2} & 2024 & \textsf{data} & \textsf{IL} & \dsim & \textsf{arm} & \nn\\
\hspace{0.6em}\textbf{RoboDreamer}~\cite{robodreamer} & 2024 & \textsf{world-model} & \textsf{model-based} & \dm & \textsf{arm} & \nn\\
\addlinespace[2pt]
\multicolumn{7}{@{}l}{\rule{0pt}{2.6ex}\textbf{\textsf{Unsupervised skill discovery \& hierarchical RL}}~\textcolor{gray}{(25)}}\\[1pt]
\multicolumn{7}{@{}l}{\hspace{0.6em}\emph{Unsupervised skill discovery}}\\
\hspace{0.6em}\textbf{VIC}~\cite{vic} & 2016 & \textsf{skill} & \textsf{RL} & \dsim & \dm & \nn\\
\hspace{0.6em}\textbf{Eigenoption}~\cite{eigenoption} & 2018 & \textsf{skill} & \textsf{RL} & \dgame & \dm & \nn\\
\hspace{0.6em}\textbf{VALOR}~\cite{valor} & 2018 & \textsf{skill} & \textsf{RL} & \dsim & \dm & \nn\\
\hspace{0.6em}\textbf{EDL}~\cite{edl} & 2020 & \textsf{skill} & \textsf{RL} & \dsim & \dm & \nn\\
\hspace{0.6em}\textbf{VISR}~\cite{visr} & 2020 & \textsf{skill} & \textsf{RL} & \dgame & \dm & \pp\\
\hspace{0.6em}\textbf{APS}~\cite{aps} & 2021 & \textsf{skill} & \textsf{RL} & \dgame & \dm & \nn\\
\hspace{0.6em}\textbf{APT}~\cite{apt} & 2021 & \textsf{skill} & \textsf{RL} & \dgame & \dm & \nn\\
\hspace{0.6em}\textbf{ProtoRL}~\cite{protorl} & 2021 & \textsf{repr} & \textsf{self-sup} & \dsim & \dm & \nn\\
\hspace{0.6em}\textbf{WURL}~\cite{wurl} & 2022 & \textsf{skill} & \textsf{RL} & \dsim & \dm & \nn\\
\hspace{0.6em}\textbf{BeCL}~\cite{becl} & 2023 & \textsf{skill} & \textsf{RL} & \dsim & \dm & \nn\\
\hspace{0.6em}\textbf{Choreographer}~\cite{choreographer} & 2023 & \textsf{skill} & \textsf{model-based} & \dsim & \dm & \nn\\
\hspace{0.6em}\textbf{CSD}~\cite{csd} & 2023 & \textsf{skill} & \textsf{RL} & \dsim & \textsf{multi} & \nn\\
\hspace{0.6em}\textbf{DGPO}~\cite{dgpo} & 2024 & \textsf{skill} & \textsf{RL} & \dsim & \dm & \nn\\
\multicolumn{7}{@{}l}{\hspace{0.6em}\emph{Hierarchical / option RL}}\\
\hspace{0.6em}\textbf{h-DQN}~\cite{hdqn} & 2016 & \textsf{policy} & \textsf{RL} & \dgame & \dm & \nn\\
\hspace{0.6em}\textbf{FeUdal Networks}~\cite{feudalnet} & 2017 & \textsf{policy} & \textsf{RL} & \dgame & \dm & \nn\\
\hspace{0.6em}\textbf{Option-Critic}~\cite{optioncritic} & 2017 & \textsf{skill} & \textsf{RL} & \dgame & \dm & \nn\\
\hspace{0.6em}\textbf{SNN4HRL}~\cite{snn4hrl} & 2017 & \textsf{skill} & \textsf{RL} & \dsim & \textsf{multi} & \nn\\
\hspace{0.6em}\textbf{HIRO}~\cite{hiro} & 2018 & \textsf{policy} & \textsf{RL} & \dsim & \textsf{legged} & \nn\\
\hspace{0.6em}\textbf{MLSH}~\cite{mlsh} & 2018 & \textsf{skill} & \textsf{RL} & \dsim & \textsf{multi} & \pp\\
\hspace{0.6em}\textbf{HAC}~\cite{hac} & 2019 & \textsf{policy} & \textsf{RL} & \dsim & \textsf{arm} & \nn\\
\hspace{0.6em}\textbf{MCP}~\cite{mcp} & 2019 & \textsf{skill} & \textsf{RL} & \dsim & \textsf{humanoid} & \nn\\
\hspace{0.6em}\textbf{NPMP}~\cite{npmp} & 2019 & \textsf{skill} & \textsf{IL} & \dsim & \textsf{humanoid} & \nn\\
\hspace{0.6em}\textbf{Relay Policy Learning}~\cite{relaypolicy} & 2019 & \textsf{policy} & \textsf{IL} & \dsim & \textsf{arm} & \nn\\
\hspace{0.6em}\textbf{HIDIO}~\cite{hidio} & 2021 & \textsf{skill} & \textsf{RL} & \dsim & \textsf{multi} & \nn\\
\hspace{0.6em}\textbf{Director}~\cite{director} & 2022 & \textsf{policy} & \textsf{model-based} & \dsim & \dm & \nn\\
\addlinespace[2pt]
\multicolumn{7}{@{}l}{\rule{0pt}{2.6ex}\textbf{\textsf{World models \& model-based RL}}~\textcolor{gray}{(24)}}\\[1pt]
\multicolumn{7}{@{}l}{\hspace{0.6em}\emph{World models}}\\
\hspace{0.6em}\textbf{World Models}~\cite{worldmodels} & 2018 & \textsf{world-model} & \textsf{model-based} & \dgame & \dm & \nn\\
\hspace{0.6em}\textbf{PlaNet}~\cite{planet} & 2019 & \textsf{world-model} & \textsf{model-based} & \dsim & \dm & \nn\\
\hspace{0.6em}\textbf{Dreamer}~\cite{dreamer} & 2020 & \textsf{world-model} & \textsf{model-based} & \dsim & \dm & \nn\\
\hspace{0.6em}\textbf{DreamerV2}~\cite{dreamerv2} & 2021 & \textsf{world-model} & \textsf{model-based} & \dgame & \dm & \nn\\
\hspace{0.6em}\textbf{DayDreamer}~\cite{daydreamer} & 2022 & \textsf{world-model} & \textsf{model-based} & \dreal & \textsf{multi} & \yy\\
\hspace{0.6em}\textbf{Iso-Dream}~\cite{isodream} & 2022 & \textsf{world-model} & \textsf{model-based} & \dsim & \dm & \nn\\
\hspace{0.6em}\textbf{MWM}~\cite{mwm} & 2022 & \textsf{world-model} & \textsf{model-based} & \dsim & \textsf{arm} & \nn\\
\hspace{0.6em}\textbf{TransDreamer}~\cite{transdreamer} & 2022 & \textsf{world-model} & \textsf{model-based} & \dgame & \dm & \nn\\
\hspace{0.6em}\textbf{DreamerV3}~\cite{dreamerv3} & 2023 & \textsf{world-model} & \textsf{model-based} & \dgame & \dm & \nn\\
\hspace{0.6em}\textbf{IRIS}~\cite{iris} & 2023 & \textsf{world-model} & \textsf{model-based} & \dgame & \dm & \nn\\
\hspace{0.6em}\textbf{STORM}~\cite{storm} & 2023 & \textsf{world-model} & \textsf{model-based} & \dgame & \dm & \nn\\
\hspace{0.6em}\textbf{SWIM}~\cite{swim} & 2023 & \textsf{world-model} & \textsf{model-based} & \dreal & \textsf{arm} & \nn\\
\hspace{0.6em}\textbf{TWM}~\cite{twm} & 2023 & \textsf{world-model} & \textsf{model-based} & \dgame & \dm & \nn\\
\hspace{0.6em}\textbf{DIAMOND}~\cite{diamond} & 2024 & \textsf{world-model} & \textsf{model-based} & \dgame & \dm & \nn\\
\hspace{0.6em}\textbf{Genie}~\cite{genie} & 2024 & \textsf{world-model} & \textsf{self-sup} & \dgame & \dm & \nn\\
\hspace{0.6em}\textbf{UniSim}~\cite{unisim} & 2024 & \textsf{world-model} & \textsf{self-sup} & \dreal & \textsf{arm} & \nn\\
\multicolumn{7}{@{}l}{\hspace{0.6em}\emph{Model-based RL}}\\
\hspace{0.6em}\textbf{PETS}~\cite{pets} & 2018 & \textsf{world-model} & \textsf{model-based} & \dsim & \dm & \nn\\
\hspace{0.6em}\textbf{MBPO}~\cite{mbpo} & 2019 & \textsf{policy} & \textsf{model-based} & \dsim & \dm & \nn\\
\hspace{0.6em}\textbf{MuZero}~\cite{muzero} & 2020 & \textsf{world-model} & \textsf{model-based} & \dgame & \dm & \nn\\
\hspace{0.6em}\textbf{SimPLe}~\cite{simple} & 2020 & \textsf{world-model} & \textsf{model-based} & \dgame & \dm & \nn\\
\hspace{0.6em}\textbf{SLAC}~\cite{slac} & 2020 & \textsf{repr} & \textsf{RL} & \dsim & \dm & \nn\\
\hspace{0.6em}\textbf{EfficientZero}~\cite{efficientzero} & 2021 & \textsf{world-model} & \textsf{model-based} & \dgame & \dm & \nn\\
\hspace{0.6em}\textbf{TD-MPC}~\cite{tdmpc} & 2022 & \textsf{world-model} & \textsf{model-based} & \dsim & \dm & \nn\\
\hspace{0.6em}\textbf{TD-MPC2}~\cite{tdmpc2} & 2024 & \textsf{world-model} & \textsf{model-based} & \dsim & \textsf{multi} & \nn\\
\addlinespace[2pt]
\multicolumn{7}{@{}l}{\rule{0pt}{2.6ex}\textbf{\textsf{Representation learning \& offline RL}}~\textcolor{gray}{(17)}}\\[1pt]
\multicolumn{7}{@{}l}{\hspace{0.6em}\emph{Visual representation learning}}\\
\hspace{0.6em}\textbf{MVP}~\cite{mvp} & 2022 & \textsf{repr} & \textsf{self-sup} & \dsim & \textsf{arm} & \nn\\
\hspace{0.6em}\textbf{PVR}~\cite{pvr} & 2022 & \textsf{repr} & \textsf{self-sup} & \dsim & \textsf{multi} & \nn\\
\hspace{0.6em}\textbf{R3M}~\cite{r3m} & 2022 & \textsf{repr} & \textsf{self-sup} & \dsim & \textsf{arm} & \nn\\
\hspace{0.6em}\textbf{LIV}~\cite{liv} & 2023 & \textsf{reward} & \textsf{self-sup} & \dreal & \textsf{arm} & \nn\\
\hspace{0.6em}\textbf{VC-1}~\cite{vc1} & 2023 & \textsf{repr} & \textsf{self-sup} & \dsim & \textsf{multi} & \nn\\
\hspace{0.6em}\textbf{VIP}~\cite{vip} & 2023 & \textsf{reward} & \textsf{self-sup} & \dreal & \textsf{arm} & \nn\\
\hspace{0.6em}\textbf{Voltron}~\cite{voltron} & 2023 & \textsf{repr} & \textsf{self-sup} & \dsim & \textsf{arm} & \nn\\
\hspace{0.6em}\textbf{Theia}~\cite{theia} & 2024 & \textsf{repr} & \textsf{self-sup} & \dsim & \textsf{arm} & \nn\\
\multicolumn{7}{@{}l}{\hspace{0.6em}\emph{Offline RL}}\\
\hspace{0.6em}\textbf{BCQ}~\cite{bcq} & 2019 & \textsf{policy} & \textsf{offline-RL} & \textcolor{gray!70}{\textsf{offline}} & \dm & \nn\\
\hspace{0.6em}\textbf{AWAC}~\cite{awac} & 2020 & \textsf{policy} & \textsf{offline-RL} & \textcolor{gray!70}{\textsf{offline}} & \dm & \pp\\
\hspace{0.6em}\textbf{CQL}~\cite{cql} & 2020 & \textsf{policy} & \textsf{offline-RL} & \textcolor{gray!70}{\textsf{offline}} & \dm & \nn\\
\hspace{0.6em}\textbf{Decision Transformer}~\cite{decisiontransformer} & 2021 & \textsf{policy} & \textsf{offline-RL} & \textcolor{gray!70}{\textsf{offline}} & \dm & \nn\\
\hspace{0.6em}\textbf{TD3+BC}~\cite{td3bc} & 2021 & \textsf{policy} & \textsf{offline-RL} & \textcolor{gray!70}{\textsf{offline}} & \dm & \nn\\
\hspace{0.6em}\textbf{Trajectory Transformer}~\cite{trajectorytransformer} & 2021 & \textsf{policy} & \textsf{offline-RL} & \textcolor{gray!70}{\textsf{offline}} & \dm & \nn\\
\hspace{0.6em}\textbf{IQL}~\cite{iql} & 2022 & \textsf{policy} & \textsf{offline-RL} & \textcolor{gray!70}{\textsf{offline}} & \dm & \nn\\
\hspace{0.6em}\textbf{Cal-QL}~\cite{calql} & 2023 & \textsf{policy} & \textsf{offline-RL} & \textcolor{gray!70}{\textsf{offline}} & \dm & \pp\\
\hspace{0.6em}\textbf{Diffusion-QL}~\cite{diffusionql} & 2023 & \textsf{policy} & \textsf{offline-RL} & \textcolor{gray!70}{\textsf{offline}} & \dm & \nn\\
\addlinespace[2pt]
\multicolumn{7}{@{}l}{\rule{0pt}{2.6ex}\textbf{\textsf{Manipulation, dexterity \& sim-to-real}}~\textcolor{gray}{(16)}}\\[1pt]
\multicolumn{7}{@{}l}{\hspace{0.6em}\emph{Manipulation / grasping}}\\
\hspace{0.6em}\textbf{Dex-Net 2.0}~\cite{dexnet2} & 2017 & \textsf{policy} & \dm & \dreal & \textsf{arm} & \nn\\
\hspace{0.6em}\textbf{GraspNet-1Billion}~\cite{graspnet1b} & 2020 & \textsf{bench} & \dm & \dreal & \textsf{arm} & \nn\\
\hspace{0.6em}\textbf{Transporter Nets}~\cite{transporter} & 2020 & \textsf{policy} & \textsf{IL} & \dsim & \textsf{arm} & \nn\\
\hspace{0.6em}\textbf{CLIPort}~\cite{cliport} & 2021 & \textsf{policy} & \textsf{IL} & \dsim & \textsf{arm} & \nn\\
\hspace{0.6em}\textbf{Contact-GraspNet}~\cite{contactgraspnet} & 2021 & \textsf{code} & \dm & \dreal & \textsf{arm} & \nn\\
\hspace{0.6em}\textbf{AnyGrasp}~\cite{anygrasp} & 2023 & \textsf{code} & \dm & \dreal & \textsf{arm} & \nn\\
\multicolumn{7}{@{}l}{\hspace{0.6em}\emph{Dexterous / in-hand}}\\
\hspace{0.6em}\textbf{Dactyl}~\cite{dactyl} & 2018 & \textsf{policy} & \textsf{RL} & \dreal & \textsf{hand} & \pp\\
\hspace{0.6em}\textbf{DAPG}~\cite{dapg} & 2018 & \textsf{policy} & \textsf{RL} & \dsim & \textsf{hand} & \nn\\
\hspace{0.6em}\textbf{Rubiks Cube Hand}~\cite{rubikscube} & 2019 & \textsf{policy} & \textsf{RL} & \dreal & \textsf{hand} & \pp\\
\hspace{0.6em}\textbf{DexPilot}~\cite{dexpilot} & 2020 & \textsf{code} & \dm & \dreal & \textsf{hand} & \nn\\
\hspace{0.6em}\textbf{Visual Dexterity}~\cite{visualdexterity} & 2023 & \textsf{policy} & \textsf{RL} & \dreal & \textsf{hand} & \nn\\
\multicolumn{7}{@{}l}{\hspace{0.6em}\emph{Sim-to-real transfer}}\\
\hspace{0.6em}\textbf{CAD2RL}~\cite{cad2rl} & 2017 & \textsf{policy} & \textsf{RL} & \dreal & \textsf{mobile} & \nn\\
\hspace{0.6em}\textbf{Domain Randomization}~\cite{domainrand} & 2017 & \textsf{code} & \dm & \dreal & \textsf{arm} & \nn\\
\hspace{0.6em}\textbf{Dynamics Randomization}~\cite{dynrand} & 2018 & \textsf{policy} & \textsf{RL} & \dreal & \textsf{arm} & \pp\\
\hspace{0.6em}\textbf{RCAN}~\cite{rcan} & 2019 & \textsf{policy} & \textsf{RL} & \dreal & \textsf{arm} & \nn\\
\hspace{0.6em}\textbf{SimOpt}~\cite{simopt} & 2019 & \textsf{policy} & \textsf{RL} & \dreal & \textsf{arm} & \pp\\
\addlinespace[2pt]
\multicolumn{7}{@{}l}{\rule{0pt}{2.6ex}\textbf{\textsf{Locomotion \& humanoid control}}~\textcolor{gray}{(16)}}\\[1pt]
\multicolumn{7}{@{}l}{\hspace{0.6em}\emph{Legged locomotion}}\\
\hspace{0.6em}\textbf{ANYmal}~\cite{anymal} & 2019 & \textsf{policy} & \textsf{RL} & \dreal & \textsf{legged} & \nn\\
\hspace{0.6em}\textbf{Quadruped Terrain}~\cite{quadterrain} & 2020 & \textsf{policy} & \textsf{RL} & \dreal & \textsf{legged} & \pp\\
\hspace{0.6em}\textbf{Legged Gym}~\cite{walkminutes} & 2021 & \textsf{code} & \textsf{RL} & \dreal & \textsf{legged} & \nn\\
\hspace{0.6em}\textbf{RMA}~\cite{rma} & 2021 & \textsf{policy} & \textsf{RL} & \dreal & \textsf{legged} & \pp\\
\hspace{0.6em}\textbf{Egocentric Locomotion}~\cite{egoloco} & 2022 & \textsf{policy} & \textsf{RL} & \dreal & \textsf{legged} & \pp\\
\hspace{0.6em}\textbf{Perceptive Locomotion}~\cite{perceptiveloco} & 2022 & \textsf{policy} & \textsf{RL} & \dreal & \textsf{legged} & \pp\\
\hspace{0.6em}\textbf{Walk These Ways}~\cite{walktheseways} & 2022 & \textsf{policy} & \textsf{RL} & \dreal & \textsf{legged} & \nn\\
\hspace{0.6em}\textbf{Extreme Parkour}~\cite{extremeparkour} & 2024 & \textsf{policy} & \textsf{RL} & \dreal & \textsf{legged} & \nn\\
\multicolumn{7}{@{}l}{\hspace{0.6em}\emph{Humanoid control}}\\
\hspace{0.6em}\textbf{ExBody}~\cite{exbody} & 2024 & \textsf{policy} & \textsf{RL} & \dreal & \textsf{humanoid} & \nn\\
\hspace{0.6em}\textbf{H2O}~\cite{h2o} & 2024 & \textsf{policy} & \textsf{RL} & \dreal & \textsf{humanoid} & \nn\\
\hspace{0.6em}\textbf{Humanoid Locomotion}~\cite{humanoidloco} & 2024 & \textsf{policy} & \textsf{RL} & \dreal & \textsf{humanoid} & \pp\\
\hspace{0.6em}\textbf{Humanoid Parkour}~\cite{humanoidparkour} & 2024 & \textsf{policy} & \textsf{RL} & \dreal & \textsf{humanoid} & \nn\\
\hspace{0.6em}\textbf{HumanPlus}~\cite{humanplus} & 2024 & \textsf{policy} & \textsf{IL} & \dreal & \textsf{humanoid} & \nn\\
\hspace{0.6em}\textbf{OmniH2O}~\cite{omnih2o} & 2024 & \textsf{policy} & \textsf{RL} & \dreal & \textsf{humanoid} & \nn\\
\hspace{0.6em}\textbf{ASAP}~\cite{asap} & 2025 & \textsf{policy} & \textsf{RL} & \dreal & \textsf{humanoid} & \pp\\
\hspace{0.6em}\textbf{HOVER}~\cite{hover} & 2025 & \textsf{policy} & \textsf{RL} & \dreal & \textsf{humanoid} & \nn\\
\addlinespace[2pt]
\multicolumn{7}{@{}l}{\rule{0pt}{2.6ex}\textbf{\textsf{Navigation, tactile \& affordance}}~\textcolor{gray}{(15)}}\\[1pt]
\multicolumn{7}{@{}l}{\hspace{0.6em}\emph{Visual / language navigation}}\\
\hspace{0.6em}\textbf{LM-Nav}~\cite{lmnav} & 2022 & \textsf{plan} & \textsf{LLM} & \dreal & \textsf{mobile} & \nn\\
\hspace{0.6em}\textbf{ViKiNG}~\cite{viking} & 2022 & \textsf{policy} & \textsf{IL} & \dreal & \textsf{mobile} & \nn\\
\hspace{0.6em}\textbf{CoW}~\cite{cow} & 2023 & \textsf{bench} & \dm & \dsim & \textsf{mobile} & \nn\\
\hspace{0.6em}\textbf{GNM}~\cite{gnm} & 2023 & \textsf{policy} & \textsf{IL} & \dreal & \textsf{multi} & \nn\\
\hspace{0.6em}\textbf{ViNT}~\cite{vint} & 2023 & \textsf{policy} & \textsf{IL} & \dreal & \textsf{mobile} & \nn\\
\hspace{0.6em}\textbf{NoMaD}~\cite{nomad} & 2024 & \textsf{policy} & \textsf{IL} & \dreal & \textsf{mobile} & \nn\\
\hspace{0.6em}\textbf{VLFM}~\cite{vlfm} & 2024 & \textsf{policy} & \textsf{LLM} & \dsim & \textsf{mobile} & \nn\\
\multicolumn{7}{@{}l}{\hspace{0.6em}\emph{Tactile \& affordance}}\\
\hspace{0.6em}\textbf{DIGIT}~\cite{digit} & 2020 & \dm & \dm & \dreal & \textsf{hand} & \nn\\
\hspace{0.6em}\textbf{Where2Act}~\cite{where2act} & 2021 & \textsf{repr} & \textsf{self-sup} & \dsim & \textsf{arm} & \nn\\
\hspace{0.6em}\textbf{FlowBot3D}~\cite{flowbot3d} & 2022 & \textsf{policy} & \textsf{self-sup} & \dsim & \textsf{arm} & \nn\\
\hspace{0.6em}\textbf{HULC}~\cite{hulc} & 2022 & \textsf{policy} & \textsf{IL} & \dsim & \textsf{arm} & \nn\\
\hspace{0.6em}\textbf{VAT-Mart}~\cite{vatmart} & 2022 & \textsf{plan} & \textsf{RL} & \dsim & \textsf{arm} & \nn\\
\hspace{0.6em}\textbf{See-to-Touch}~\cite{seetotouch} & 2023 & \textsf{policy} & \textsf{RL} & \dreal & \textsf{hand} & \pp\\
\hspace{0.6em}\textbf{T-DEX}~\cite{tdex} & 2023 & \textsf{repr} & \textsf{self-sup} & \dreal & \textsf{hand} & \nn\\
\hspace{0.6em}\textbf{VRB}~\cite{vrb} & 2023 & \textsf{repr} & \textsf{self-sup} & \dreal & \textsf{arm} & \nn\\
\addlinespace[2pt]
\multicolumn{7}{@{}l}{\rule{0pt}{2.6ex}\textbf{\textsf{Datasets, benchmarks \& simulators}}~\textcolor{gray}{(16)}}\\[1pt]
\multicolumn{7}{@{}l}{\hspace{0.6em}\emph{Datasets}}\\
\hspace{0.6em}\textbf{RoboNet}~\cite{robonet} & 2019 & \textsf{data} & \textsf{model-based} & \dreal & \textsf{arm} & \nn\\
\hspace{0.6em}\textbf{Ego4D}~\cite{ego4d} & 2022 & \textsf{data} & \dm & \textcolor{gray!70}{\textsf{offline}} & \dm & \nn\\
\hspace{0.6em}\textbf{BridgeData V2}~\cite{bridgev2} & 2023 & \textsf{data} & \textsf{IL} & \dreal & \textsf{arm} & \nn\\
\hspace{0.6em}\textbf{DobbE}~\cite{dobbe} & 2023 & \textsf{data} & \textsf{IL} & \dreal & \textsf{mobile} & \nn\\
\hspace{0.6em}\textbf{RH20T}~\cite{rh20t} & 2023 & \textsf{data} & \textsf{IL} & \dreal & \textsf{arm} & \nn\\
\hspace{0.6em}\textbf{DROID}~\cite{droid} & 2024 & \textsf{data} & \textsf{IL} & \dreal & \textsf{arm} & \nn\\
\multicolumn{7}{@{}l}{\hspace{0.6em}\emph{Benchmarks}}\\
\hspace{0.6em}\textbf{D4RL}~\cite{d4rl} & 2020 & \textsf{bench} & \textsf{offline-RL} & \textcolor{gray!70}{\textsf{offline}} & \dm & \nn\\
\hspace{0.6em}\textbf{ManiSkill}~\cite{maniskill} & 2021 & \textsf{bench} & \textsf{IL} & \dsim & \textsf{arm} & \nn\\
\multicolumn{7}{@{}l}{\hspace{0.6em}\emph{Simulators}}\\
\hspace{0.6em}\textbf{AI2-THOR}~\cite{ai2thor} & 2017 & \textsf{code} & \dm & \dsim & \textsf{mobile} & \nn\\
\hspace{0.6em}\textbf{Habitat}~\cite{habitat} & 2019 & \textsf{code} & \dm & \dsim & \textsf{mobile} & \nn\\
\hspace{0.6em}\textbf{SAPIEN}~\cite{sapien} & 2020 & \textsf{code} & \dm & \dsim & \textsf{arm} & \nn\\
\hspace{0.6em}\textbf{ThreeDWorld}~\cite{tdw} & 2020 & \textsf{code} & \dm & \dsim & \textsf{multi} & \nn\\
\hspace{0.6em}\textbf{iGibson}~\cite{igibson} & 2021 & \textsf{code} & \dm & \dsim & \textsf{mobile} & \nn\\
\hspace{0.6em}\textbf{Isaac Gym}~\cite{isaacgym} & 2021 & \textsf{code} & \dm & \dsim & \textsf{multi} & \nn\\
\hspace{0.6em}\textbf{Isaac Lab Orbit}~\cite{orbit} & 2023 & \textsf{code} & \dm & \dsim & \textsf{multi} & \nn\\
\hspace{0.6em}\textbf{RoboCasa}~\cite{robocasa} & 2024 & \textsf{bench} & \textsf{IL} & \dsim & \textsf{mobile} & \nn\\
\addlinespace[2pt]
\end{longtable}
}

\end{document}